\documentclass[lettersize,journal]{IEEEtran}
\usepackage{algorithmic}
\usepackage{array}
\usepackage[caption=false,font=normalsize,labelfont=sf,textfont=sf]{subfig}
\usepackage{textcomp}
\usepackage{stfloats}
\usepackage{url}
\usepackage{verbatim}
\usepackage{graphicx}
\usepackage{rotating}  %

\usepackage{multirow,bbding}
\usepackage{amsmath,amssymb,amsfonts,bm}
\usepackage{threeparttable}

\usepackage{booktabs}

\usepackage{tcolorbox}
\usepackage{mdframed}

\usepackage{wasysym}

\usepackage{cite}
\usepackage{orcidlink} %

\usepackage{tablefootnote}

\usepackage{longtable}

\usepackage{hyperref}
\hypersetup{
hidelinks
}

\usepackage{fontawesome5}

\usepackage{pifont}

\newcommand{\cmark}{\textcolor{green!70!black}{\ding{51}}} %
\newcommand{\xmark}{\textcolor{red}{\ding{55}}}  

\usepackage{tikz}%
\definecolor{lime}{HTML}{A6CE39}
\DeclareRobustCommand{\orcidicon}{%
\begin{tikzpicture}
\draw[lime, fill=lime] (0,0) 
circle [radius=0.16] 
node[white] {{\fontfamily{qag}\selectfont \tiny ID}};\draw[white, fill=white] (-0.0625,0.095) 
circle [radius=0.007];\end{tikzpicture}
\hspace{-2mm}}
\foreach \x in {A, ..., Z}{%
\expandafter\xdef\csname orcid\x\endcsname{\noexpand\href{https://orcid.org/\csname orcidauthor\x\endcsname}{\noexpand\orcidicon}}
}

\def\BibTeX{{\rm B\kern-.05em{\sc i\kern-.025em b}\kern-.08em
T\kern-.1667em\lower.7ex\hbox{E}\kern-.125emX}}
\usepackage{balance}
\begin{document}

\tikzstyle{box} = [rectangle, minimum width=3cm, minimum height=1cm, text centered, draw=black]
\tikzstyle{arrow} = [thick,->,>=stealth]

\usetikzlibrary{positioning}

\title{CalFuse: Multi-Modal Continual Learning via Feature Calibration and Parameter Fusion}

\author{
\centering
Juncen Guo\orcidlink{0009-0004-0345-1787}, 
Siao Liu, 
Xiaogunag Zhu\orcidlink{0000-0001-9554-2133}, 
Lianlong Sun\orcidlink{0009-0005-3208-9075}, 
Liangyu Teng\orcidlink{0009-0003-3166-4977}, 
Jingyi Wu\orcidlink{0000-0001-8283-0646}, 
Di Li\orcidlink{0000-0002-7764-7609}, 
Linxiao Gong\orcidlink{0009-0003-0112-8305},
Weiwei Jiang,
Wei Zhou\orcidlink{0000-0003-3641-1429}, ~\textit{Senior Member, IEEE},
Liang Song\orcidlink{0000-0002-8143-9052}~\textit{Senior Member, IEEE}

\thanks{
  This work was supported by National Key Research and Development Program of China, Project No.2024YFE0200700, Subject No.2024YFE0200703. This work was also supported by China State Construction Engineering (HONG KONG) Limited. And this work was also supported in part by the Specific Research Fund of the Innovation Platform for Academicians of Hainan Province under Grant YSPTZX202314, in part by the Shanghai Key Research Laboratory of NSAI and the China FAW Joint Development Project. \textit{(Corresponding author: Liang Song.)}
}

\thanks{Juncen Guo, Liangyu Teng, Jingyi Wu, and Liang Song are with the College of Intelligent Robotics and Advanced Manufacturing, Fudan University, Shanghai, 200433, China (email: guojc23@m.fudan.edu.cn;  lyteng20@fudan.edu.cn; jingyiwu23@m.fudan.edu.cn; songl@fudan.edu.cn).}

\thanks{Siao Liu is with the School of Future Science and Engineering, Soochow University, China  (email: saliu@suda.edu.cn). Xiaoguang Zhu is with the DataLab: Data Science and Informatics, University of California, Davis, CA, 95616, USA (email: xgzhu@ucdavis.edu). Lianlong Sun is with the University of Rochester, New York, 14627, USA (email: lianlongsun@rochester.edu). Di Li is with the Ningbo University, China (email: lidi1@nbu.edu.cn). Linxiao Gong is with the Hong Kong University of Science and Technology (Guangzhou), China (email: lgong265@connect.hkust-gz.edu.cn). Weiwei Jiang is with the Beijing University of Posts and Telecommunications, China (email: jww@bupt.edu.cn). Wei Zhou is with the Academy for Computer Science and Informatics, Cardiff University, Wales, CF24 4AG, UK (email: zhouw26@cardiff.ac.uk).}

}

\markboth{IEEE Transactions on Big Data,~Vol.~xx, No.~xx, xxxx~xxxx}%
{CalFuse: Feature Calibration Enhanced Parameter Fusion for Class-Continual Learning}

\maketitle

\begin{abstract}
With the proliferation of multi-modal data in large-scale visual recognition systems, enabling models to continuously acquire knowledge from evolving data streams while preserving prior information has become increasingly critical. Class-Continual Learning (CCL) addresses this challenge by incrementally incorporating new class knowledge without revisiting historical data, making it essential for real-world big data applications. While traditional CCL methods rely solely on visual features, recent advances in Vision-Language Models (VLMs) such as CLIP demonstrate significant potential for CCL by leveraging pre-trained multi-modal knowledge. However, existing approaches face challenges in mitigating catastrophic forgetting while maintaining the cross-modal generalization capabilities of VLMs. To address these limitations, we propose \textbf{CalFuse}, a framework that synergizes feature \underline{Cal}ibration with parameter \underline{Fus}ion to enable effective multi-modal knowledge integration in continual learning scenarios. CalFuse introduces a dynamic feature calibration mechanism that adaptively balances original CLIP visual representations with task-specific features, preserving the model's intrinsic cross-modal generalization while adapting to new classes. Concurrently, a QR decomposition-based parameter fusion strategy progressively integrates newly acquired knowledge with historical task parameters, maintaining equilibrium between learning new class representations and retaining prior knowledge across sequential tasks. Extensive experiments on benchmark datasets  validate the effectiveness of our approach in large-scale multi-modal continual learning settings, demonstrating superior performance over state-of-the-art methods in both average accuracy and final task retention.
\end{abstract}


\section{Introduction}
\label{sec:intro}

\IEEEPARstart{I}{n} the era of big data, the ubiquity of large-scale multi-modal datasets presents both opportunities and challenges for developing intelligent systems capable of continuous adaptation. Modern visual recognition applications, ranging from autonomous vehicles to medical imaging systems, must process massive volumes of data arriving sequentially while maintaining knowledge of previously learned concepts. Class-Continual Learning (CCL) emerges as a fundamental paradigm to address this challenge, enabling models to incrementally acquire knowledge from new classes while retaining information from former tasks \cite{p87,p88,p89}. Moreover, CCL is particularly crucial for real-world deployments \cite{p90,p91,p69,p86,p73,p75,p84,p85} where complete retraining is prohibitively expensive and historical data may be inaccessible due to privacy constraints or storage limitations. Furthermore, the significance of CCL is exemplified in dynamic environments where data distributions evolve continuously. For instance, autonomous driving systems must adapt to emerging traffic patterns, evolving road infrastructure, and changing weather conditions \cite{p61,p71}, while medical imaging platforms face the challenge of recognizing newly discovered diseases and evolving pathological presentations. These scenarios demand CCL-enabled models that can effectively integrate new information without catastrophic forgetting—a necessity amplified by the constraints of edge devices with limited computational capacity and storage.

\begin{figure}[t]
  \centering
  \includegraphics[width=\linewidth]{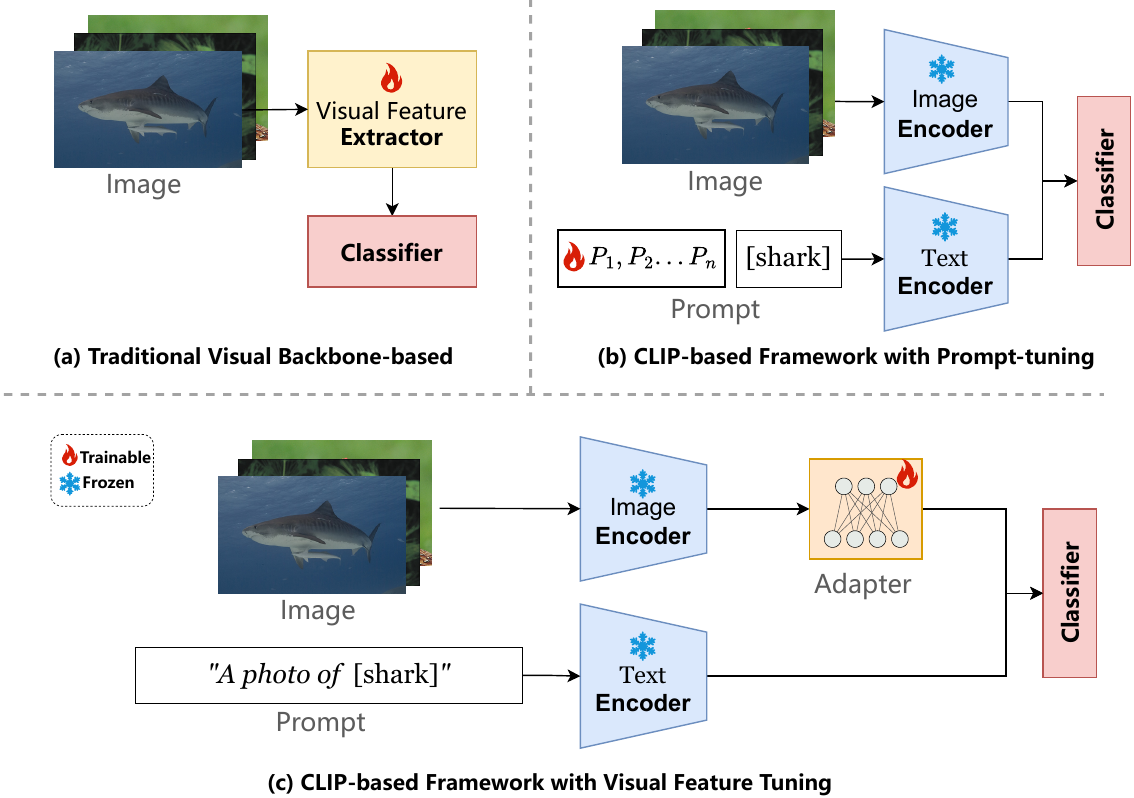}
  \caption{Illustration of existing CCL frameworks. (a) Traditional vision-only backbone-based CCL. (b) CLIP-based framework with prompt-tuning focusing on text modality. (c) CLIP-based framework with visual feature adaptation. Our approach addresses limitations in both paradigms by preserving multi-modal representations while enabling effective continual adaptation.}
  \label{fig1}
\end{figure}

A primary challenge in CCL lies in balancing \textit{stability: the retention of previously acquired knowledge}, and \textit{plasticity: the adaptability to accommodate new tasks} \cite{p20}. To mitigate catastrophic forgetting \cite{p20}, various methodologies have been proposed, including replay-based approaches \cite{p55}, regularization techniques \cite{p29}, and parameter-isolated methods \cite{p52}. While these strategies have demonstrated success with foundational vision backbones such as ResNet \cite{p59} and Vision Transformer (ViT) \cite{p60}, as illustrated in Fig.~\ref{fig1}(a), their exclusive reliance on visual features limits their effectiveness in complex scenarios where heterogeneous multi-modal information could provide richer semantic representations.

Recent advances in multi-modal large language representation learning have revolutionized the landscape of visual recognition. Vision-Language Models (VLMs), particularly Contrastive Language-Image Pre-training (CLIP) \cite{p13}, have been pre-trained on massive-scale image-text pairs, learning unified cross-modal embeddings that bridge visual and textual semantics. These models exhibit remarkable zero-shot transfer capabilities and rich feature representations that can potentially bridge the knowledge gap between classes learned at different continual learning stages. Continual-CLIP \cite{p15} pioneered the application of frozen CLIP backbones to CCL, demonstrating that VLMs' pre-trained multi-modal knowledge is inherently suited for continual scenarios. Subsequent methods have explored various fine-tuning strategies \cite{p18}, consistently outperforming traditional vision-only approaches.

\begin{figure}
  \centering
  \includegraphics[width=1\linewidth]{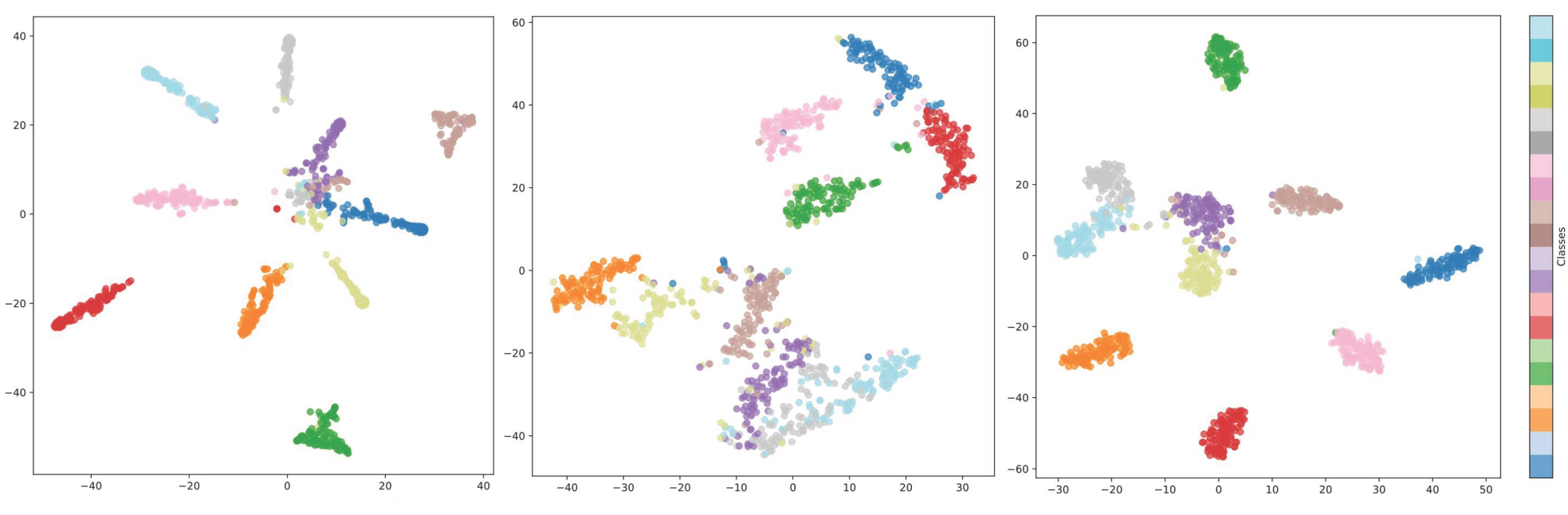}
  \caption{T-SNE visualization comparing feature space quality on CIFAR100 B0 Inc10 after the second task: (a) Continual-CLIP (frozen CLIP features), (b) visual feature tuning approach showing degraded cluster separation, and (c) our CalFuse method maintaining well-separated clusters while adapting to new tasks.}
  \label{figx}
\end{figure}

Current CLIP-based CCL methods primarily fall into two categories: prompt-tuning \cite{p16,p17} and visual feature tuning \cite{p50}, as shown in Fig.~\ref{fig1}(b) and (c). Prompt-tuning methods optimize task-specific or input-conditioned text prompts while keeping the backbone frozen, focusing predominantly on the textual modality. Visual feature tuning methods, conversely, introduce trainable adaptation layers to transform CLIP features into task-specific representations. However, both approaches face critical limitations in the context of large-scale multi-modal continual learning. First, prompt-tuning methods primarily enhance the text prompt space while underutilizing the rich visual features embedded in CLIP's pre-trained representations. This unbalanced emphasis on a single modality fails to fully exploit the cross-modal synergy that makes VLMs powerful. Second, visual feature tuning methods risk degrading CLIP's original feature extraction capabilities, as direct fine-tuning may distort the carefully learned multi-modal alignment. As evidenced in Fig.~\ref{figx}, the classification clusters produced by visual feature tuning methods exhibit reduced separability compared to zero-shot CLIP, indicating potential deterioration of the pre-trained feature space.

To address these limitations in multi-modal continual learning, we propose \textbf{CalFuse}, a framework that integrates Feature \underline{Cal}ibration and  Parameter \underline{Fus}ion to effectively leverage CLIP's pre-trained multi-modal knowledge while mitigating catastrophic forgetting. Our approach is motivated by two key insights: (1) preserving the integrity of CLIP's cross-modal representations is essential for maintaining generalization across diverse visual concepts, and (2) dynamically fusing knowledge across sequential tasks enables smooth integration of new information without abrupt disruption of prior learning.

Specifically, CalFuse employs a frozen CLIP encoder as the foundational architecture, ensuring that the core multi-modal alignment learned during pre-training remains intact. To address the potential distortion of CLIP's pre-trained knowledge caused by direct fine-tuning, we introduce a Feature Calibration (FC) mechanism that employs lightweight trainable layers to adaptively fuse original CLIP visual features with task-specific adaptations. The FC mechanism not only enhances discriminative power for current tasks but also incorporates a calibration parameter $\alpha$ to precisely control the fusion ratio between original and adapted features. This design ensures the model maintains robust cross-modal generalization while reducing overfitting to incremental tasks—a critical requirement for large-scale continual learning scenarios.
Complementing the feature-level adaptation, we propose a Parameter Fusion (PF) module that addresses the knowledge discrepancy between sequential tasks during classifier construction. Unlike methods that focus solely on textual prompt adaptation, our approach prioritizes dynamic knowledge fusion across all classes throughout the continual learning process. The PF module leverages QR decomposition to decompose adapter parameters into orthogonal and triangular components, enabling principled fusion of parameters from adjacent learning stages. As tasks evolve, parameters undergo structured decomposition and recombination, effectively alleviating catastrophic forgetting while accommodating new knowledge. Furthermore, we incorporate a dynamic distillation loss with an adaptive weighting scheme that adjusts to the changing ratio of old to new classes, providing additional regularization to constrain parameter drift.

The main contributions of this paper are summarized as follows:
\begin{itemize}
    \item We propose a multi-modal continual learning framework tailored for large-scale visual recognition, which effectively preserves CLIP's pre-trained cross-modal representations through a novel feature calibration mechanism while enabling adaptive learning of new classes.
    \item To address catastrophic forgetting in multi-modal settings, we introduce a QR decomposition-based parameter fusion strategy that captures and integrates parameter variations across sequential tasks, enabling smooth knowledge transfer and accumulation throughout the continual learning process.
    \item Extensive experiments on CIFAR100 and ImageNet100 datasets demonstrate that CalFuse achieves state-of-the-art performance under the challenging B0 setting, validating the effectiveness of our approach for multi-modal continual learning in large-scale scenarios.
\end{itemize}

The remainder of this paper is organized as follows. Section~\ref{sec2} reviews related work on CCL and VLMs, with emphasis on multi-modal representation learning. Section~\ref{sec3} introduces the overall framework of CalFuse, detailing the design of feature calibration and parameter fusion modules along with the optimization objective. Section~\ref{sec4} presents comprehensive experimental results comparing CalFuse with state-of-the-art methods on large-scale benchmarks. Finally, we conclude the paper by discussing current limitations and promising directions for future research in multi-modal continual learning systems.

\section{Related work}
\label{sec2}
\subsection{Class-Continual Learning}
CCL aims to continuously incorporate new knowledge from new data while retaining, integrating and even optimizing old knowledge to prevent catastrophic forgetting \cite{p20} in the process of continuously building a classifier for all classes \cite{p74}
. The current mainstream methods for CCL are divided into three main categories. The replay-based methods either save a portion of samples from previous tasks %
or generate synthetic samples with a generative model %
in a cache \cite{p55}. When training on new tasks, these samples are used alongside new samples to prevent forgetting previously learned knowledge. \cite{p22} selects samples with high predictive entropy and proximity to the decision boundary, aiming to enhance generalization by replaying these challenging examples. \cite{p21} estimates sample uncertainty through data augmentation, identifying instances with high prediction variance across multiple augmented versions. Regularization-based methods impose constraints on network parameter updates. Weight regularization focuses on preventing drift in weights associated with previous tasks \cite{p29}%
, while data regularization methods transfer knowledge from the old model to the new one through knowledge distillation \cite{p31}%
. Given that each parameter does not contribute equally to the task, parameter regularization methods seek to assess the importance of each parameter to the network and keep important parameters static to preserve previous knowledge. \cite{p68} examines the performance and stability of various network layers in CCL, revealing distinct characteristics. Shallow layers learn quickly but offer limited representational capacity, whereas deeper layers adapt more slowly yet provide stronger discriminative features. Knowledge distillation refers to building a mapping between a set of methods, which has been widely adopted in CCL in several ways. AFC \cite{p67} considers the importance of different feature maps for distillation.
COIL \cite{p31} suggests the use of collaborative transport for bidirectional distillation, where both the semantic relationships between old and new models are exploited. Parameter isolation-based methods mitigate catastrophic forgetting by dynamically increasing the capacity of network. It enables the model to retain knowledge from previous tasks while expanding to learn new ones \cite{p52}%
. \cite{p52} proposes the Dynamically Expandable Representation (DER) method, which decouples feature adaptation from the classifier. For a new task, DER freezes the existing feature extractor, adds a new one, and concatenates features from all extractors for classification. %
\cite{p65} proposes a dynamic structure reorganization strategy that preserves old knowledge while learning new classes. When a new class arrives, side branches are added to the network blocks. The original weights are frozen, and the new branches are trained to learn the new class. In the field of multimodal learning, CLIP has emerged as a representative large-scale vision-languag model that effectively aligns image and text modalities within a shared embedding space. Owing to its excellent zero-shot capability across both single-task and multi-task settings \cite{%
p39}%
, recent studies \cite{p50,p16,p17} have demonstrated the potential of extending CLIP to continual learning scenarios, enabling models to incrementally acquire multimodal knowledge without catastrophic forgetting.

\subsection{Vision-Language Models}
Large-scale Vision-Language Models such as CLIP \cite{p13}, ALIGN \cite{p41}, and Florence \cite{p42} have been pretrained on massive datasets containing image-text pairs, enabling them to learn a unified embedding space that bridges the gap between visual and textual modalities. As representative paradigms of multimodal learning, these models effectively align heterogeneous information sources and capture cross-modal semantic relationships, thereby enhancing the model’s ability to understand and reason across different modalities. Their remarkable generalization and transfer capabilities make them particularly effective for zero-shot learning across a wide range of downstream tasks without the need for task-specific fine-tuning. In the context of multimodal continual learning, a growing body of research has begun to explore how to adapt these powerful pretrained models to sequential tasks while minimizing catastrophic forgetting \cite{p72}. Recent approaches leverage the inherent multimodal representations of such models to support incremental knowledge acquisition across both visual and textual domains. A recent trend involves introducing additional learnable components—such as prompts or lightweight modules—that allow the pretrained backbone to remain largely unchanged while adapting to new tasks. For instance, L2P \cite{p44} brings prompt-based learning into continual learning by maintaining a shared pool of prompts that guide the model to adapt to each task dynamically.
DualPrompt \cite{p18} proposes attaching complementary prompts to the pretrained backbone and formulates the objective as learning task-invariant and task-specific instructions. Most of these works focus on prompt adjustment, prefix adjustment and effective organization of text features but ignore the effective attention to visual features. This paper aims to fill this research gap by proposing a comprehensive solution that focuses on visual features to tune visual language models without suffering from forgetting.

\section{Methodology}
\label{sec3}
\subsection{Overview}

\begin{figure*}
  \centering
  \includegraphics[width=1\textwidth]{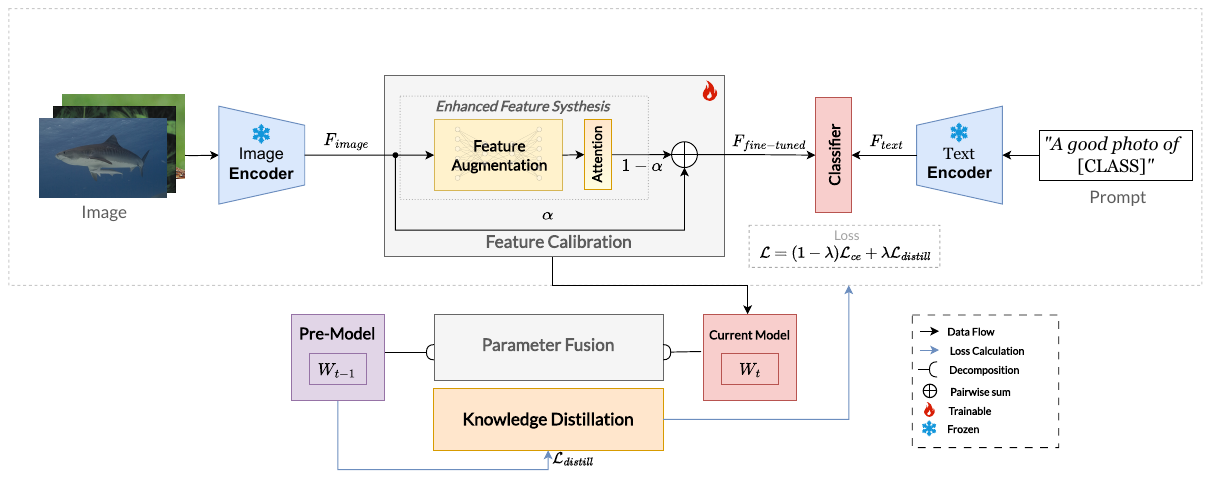}
  \caption{The framework of the proposed feature calibration enhanced parameter fusion.}
  \label{fig2}
  \end{figure*}

Traditional visual methods often struggle to make effective use of textual information, and fine-tuning methods based on CLIP have shown promising performance in multimodal settings. They are highly susceptible to catastrophic forgetting when applied to continual learning scenarios. To address these limitations, we propose a novel method called CalFuse, which effectively leverages the rich cross-modal knowledge encapsulated in CLIP while simultaneously addressing the challenge of forgetting. As shown in Fig.\ref{fig2}, CalFuse enhances performance through three core components that collectively aim to balance knowledge retention and task comprehensiveness. The first component FC is introduced to preserve and optimize the pretrained CLIP representations. To reduce the risk of distorting the pretrained CLIP knowledge caused by direct fine-tuning, FC is designed to refine the extracted visual features in a lightweight and stable manner. This module enhances the visual embeddings without altering the core CLIP encoder, ensuring that the foundational multimodal knowledge is preserved. To further maintain the generalization capability of the model while preventing overfitting to the current task, we introduce a parameter $\alpha$ after the calibration stage. This parameter determines the extent to which the original image embeddings from the CLIP Image Encoder are retained. By maintaining a proportion of these original embeddings, the model can better preserve the semantics of the pretrained representation space, leading to improved robustness across diverse tasks. 
The second core component, PF, is designed to address the issue of forgetting. It effectively fuses the current adapter parameters with those from the previous stage, serving as the initial parameters for the next stage's adapter. This progressive update mechanism allows knowledge from earlier tasks to be preserved and fused into future learning. As tasks evolve, PF accumulates experience by continuously adapting and refining its parameters, ensuring that knowledge from old classes is retained even as new ones are introduced. This design effectively mitigates catastrophic forgetting in the process of continuously building a classifier for all classes. Finally, CalFuse introduces Dynamic Distillation, a module responsible for transferring knowledge between the current and previous tasks. This component acts as a bridge between stages, enhancing continuity and semantic consistency throughout the learning process. In summary, CalFuse offers a comprehensive and systematic approach to continual learning with VLMs. By preserving cross-modal knowledge, promoting effective parameter reuse, and introducing a distillation mechanism, CalFuse achieves a strong balance between performance on new tasks and retention of previously acquired knowledge.

\subsection{Feature Calibration}
\label{sec}
Given the powerful zero-shot capabilities of CLIP and its remarkable transferability across a wide range of downstream tasks, we introduce a feature calibration mechanism to better align task-specific requirements with the pretrained CLIP representations. This mechanism is carefully designed to integrate the discriminative features required for a specific task with the generalizable and robust features extracted by CLIP, thereby enhancing the model’s adaptability without sacrificing its generalization performance. As depicted in Fig.~\ref{fig2}, the image features $\mathbf{F}_\text{fine-tuned}$ are obtained by first passing input images through the frozen CLIP image encoder, followed by a lightweight adapter module. This adapter enables subtle yet effective transformation of the visual features, adapting them to the current task while minimizing the risk of disrupting the original pretrained knowledge. In parallel, label features $\mathbf{F}_\text{text}$ are generated by processing the textual class labels through a prompt generator, and then passing them through the CLIP text encoder. The generated features capture semantic class-level information, aligned in the same embedding space as the image features. Finally, classification is performed by computing the similarity between the calibrated image features and the class text features. 

To further improve the representational capacity of the visual features while preserving compatibility with the pretrained CLIP framework, we propose a trainable adaptive module named Enhanced Feature Synthesis (EFS). This module is specifically designed to enhance feature quality through a deeper transformation process, while maintaining the original dimensionality of the input features to ensure seamless integration with downstream components. The EFS consists of four sequential linear transformation layers that enable the network to capture more complex relationships within the feature space. In addition to these transformations, the module incorporates attention mechanism techniques to dynamically recalibrate the feature responses. %
Importantly, the feature dimensionality remains unchanged before and after passing through the EFS. This design ensures that the enhanced features produced by EFS remain fully compatible with the CLIP feature space and can be directly used for similarity computation. 
To retain the original characteristics of CLIP, feature calibration is conducted as follows:
\begin{equation}
\textbf{F}_{\text {fine-tuned}}=(1-\alpha) \Phi_{EFS}\left(\textbf{F}_{\text {image }}\right)+\alpha \textbf{F}_{\text{image}},
\end{equation}
where $\alpha$ is a parameter that balances the enhanced features with the original CLIP knowledge. The first term represents the adaptive learning of knowledge from new categories in the new task, while the second term is specifically designed to retain the original knowledge from CLIP.

\subsection{Parameter Fusion}
To address the challenge of catastrophic forgetting in continual learning, we introduce parameter fusion—a mechanism that dynamically fuses adapter parameters from the current learning stage with those preserved from previous stages. It enables the progressive fusion of new and old knowledge, ensuring that the model can incrementally construct a unified classifier capable of handling all previously encountered classes. By retaining and reusing historical parameters, PF promotes long-term memory retention while simultaneously adapting to novel information, thereby mitigating the risk of performance degradation due to forgetting.

In many traditional weight update schemes, newly acquired information often overwrites past knowledge, leading to a well-known issue of catastrophic forgetting. This problem is particularly pronounced in continual learning scenarios, where the model is required to adapt continuously to a sequence of tasks without access to the full dataset from earlier stages. To further combat this issue, we propose a strategy based on QR decomposition. Specifically, during each incremental update, the model's weight matrix is decomposed into an orthogonal matrix $\mathbf{Q}$ and an upper triangular matrix $\mathbf{R}$, allowing us to restructure and recombine the parameter space in a manner that preserves essential features of previously learned tasks. The use of QR decomposition offers a mathematically grounded approach to separating and retaining knowledge components from earlier stages, while facilitating the smooth integration of new task-specific information. By recombining these components in a controlled way, the model can maintain a balanced representation across tasks, effectively reducing interference and preserving critical knowledge.

\begin{figure}
  \centering
  \includegraphics[width=.49\textwidth]{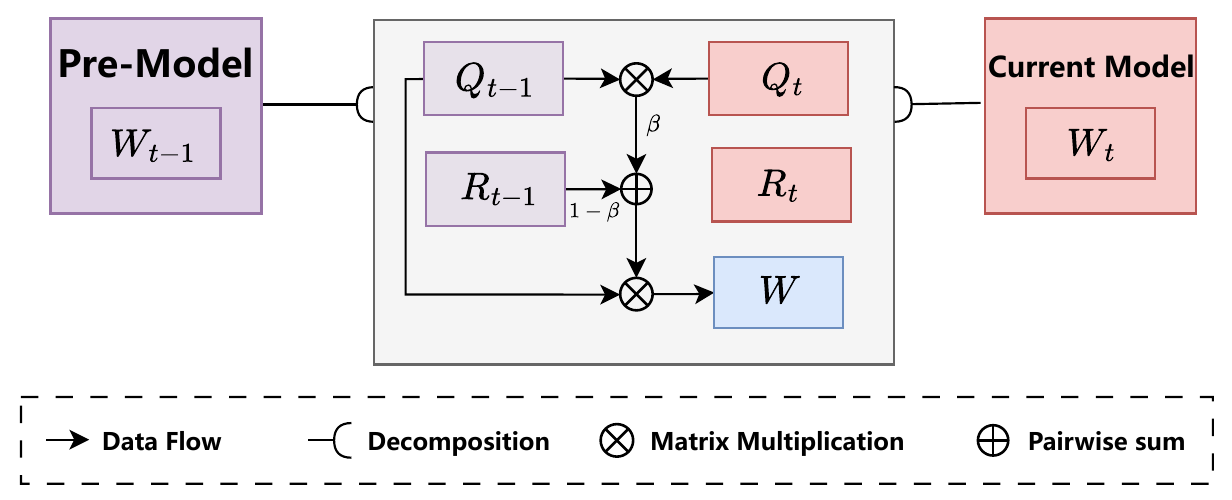}
  \caption{The parameter fusion process based on QR decomposition.}
  \label{fig3}
  \end{figure}

Specifically, QR decomposition is employed to decompose a target weight matrix into two constituent matrices: an orthogonal matrix $\mathbf{Q}$ and an upper triangular matrix $\mathbf{R}$, such that the original matrix is represented as $\mathbf{W} =\mathbf{Q} \mathbf{R}$. In this decomposition, the matrix $\mathbf{Q}$ contains orthonormal basis vectors that span the column space of $\mathbf{W}$, while $\mathbf{R}$ captures the scaling and linear combinations required to reconstruct the original representation.
As shown in Fig.\ref{fig3}, We apply QR decomposition to the old and new weight matrices 
$\{\mathbf{W}_{t-1},\mathbf{W}_{t}\}$:
\begin{equation}
\mathbf{W}_{t-1} =\mathbf{Q}_{t-1} \mathbf{R}_{t-1}, \mathbf{W}_{t}  =\mathbf{Q}_{t} \mathbf{R}_{t}.
\end{equation}

To achieve a controllable and effective integration of knowledge across tasks, we further introduce a weight coefficient $\beta$ to balance the contributions of the current and previous stages. This coefficient regulates the extent to which the newly acquired task information should influence the model, relative to the preserved knowledge from earlier tasks. Specifically, after performing QR decomposition on both the current and previous adapter weight matrices, we obtain corresponding $\mathbf{Q}$ and $\mathbf{R}$ components. We then linearly combine these components using the balancing coefficient $\beta$. The final fused weight matrix, which becomes the initial weight matrix for the model's adapter in the next stage, is updated as:
\begin{equation}
\mathbf{W}=\mathbf{Q}_{t-1} \mathbf{R}_{t + 1},
\end{equation}
\begin{equation}
\mathbf{R}_{t+1}=\left[\mathbf{Q}_{t-1}^T \otimes \mathbf{Q}_t \cdot \beta\right] \oplus\left[\mathbf{R}_{t-1} \cdot(1-\beta)\right],
\end{equation}
where $\beta$ is adjusted to maximize the retention of knowledge from both new and old tasks. $\mathbf{R}_{t+1}$ is obtained by weighted fusion of the projection of $\mathbf{Q}_{t}$ on $\mathbf{Q}_{t-1}$ with $\mathbf{R}_{t-1}$.

Through QR decomposition, we can not only avoid the parameter drift caused by direct fine-tuning but also maintain the stability of the parameters in each task iteration, thus alleviating the problem of catastrophic forgetting. The advantage of this method is that QR decomposition preserves the structural information of the old weights and effectively fuses the learning of new tasks with the knowledge from old tasks, maintaining the overall generalization ability and knowledge transfer capability of the model.

\subsection{Optimization}
\begin{figure}[t]
  \centering
  \includegraphics[width=.49\textwidth]{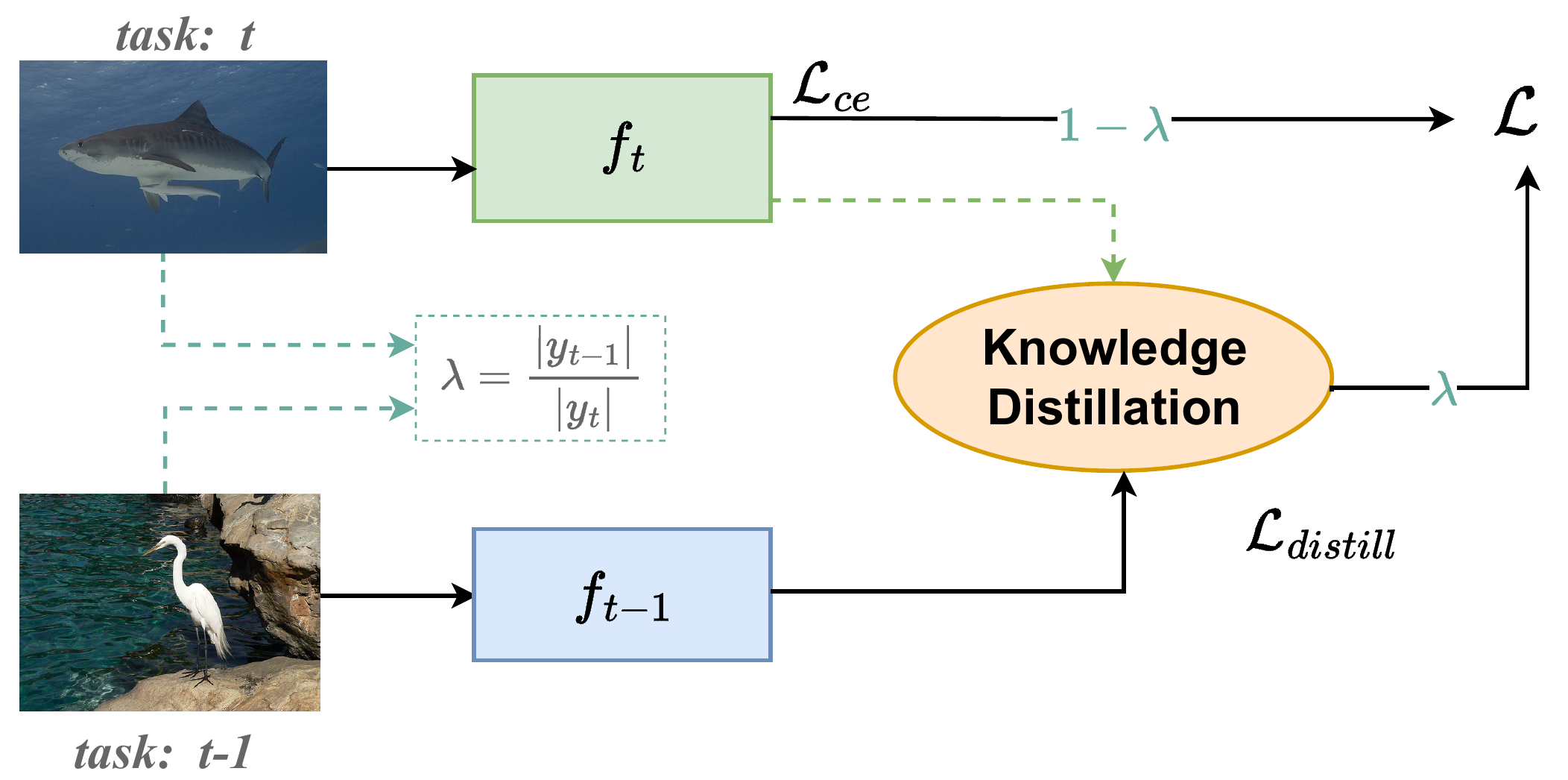}
  \caption{The loss function construction for our proposed method.}
  \label{fig5}
  \end{figure}

The loss function of our proposed method consists of two main parts: the cross-entropy loss and the dynamic distillation loss in Fig.~\ref{fig5}. 

\textbf{CLIP Loss:} CLIP consists of an image encoder and a text encoder. The image $\textbf{x}$ and the text $\textbf{z}$ are separately input into the image encoder and text encoder, respectively, to obtain $\textbf{F}_\text{image}$ and $\textbf{F}_\text{text}$ respectively. In CLIP, $\textbf{z}$ is obtained through manually crafted prompts, such as "a good photo of [ ]", where "[ ]" is replaced by the class name of the test image. Therefore, the probability of predicting the test image $\textbf{x}$ as class $y_i$ can be computed as:
\begin{equation}
p\left(y_i \mid \textbf{x}\right)=\frac{e^{<\bm{F}_{\text {image}}, \bm{F}_{\text {text}}^{y_i}>/ \tau}}{\sum_{j=1}^{y_t} e^{<\bm{F}_{\text {image}}, \bm{F}_{\text {text}}^j>/ \tau}},
\end{equation}
where $\tau$ is a temperature parameter learned by CLIP, $\langle\cdot, \cdot\rangle$ denotes the cosine similarity, $ \bm{F}_{\text {text}}^j$ is the embedding derived from $\textbf{z}_j$ of the $j$-th class, and $y_b$ is the total number of downstream dataset classes.

After the FC mechanism, we use the formula with $\bm{F}_{\text {fine-tuned}}$ to obtain the class probability vector and predict the class. The probability of predicting the test image \textbf{x} as class $y_i$ can be calculated as:
\begin{equation}
p\left(y_i \mid \textbf{x}\right)=\frac{e^{<\alpha \textbf{F}_{\text {image }}+(1-\alpha) \Phi_{EFS}\left(\textbf{F}_{\text {image }}\right), \textbf{F}_{\text {text }}^{y_i}>/ \tau}}{\sum_{j=1}^{y_t} e^{<\alpha \textbf{F}_{\text {image }}+(1-\alpha) \Phi_{EFS}(\textbf{F}_{\text {image }}), \textbf{F}_{\text {text }}^j>/ \tau}},
\end{equation}

We employ the cross-entropy loss criterion to fine-tune the parameters of the adapter, formulated as follows:
\begin{equation}
\mathcal{L}_{c e}(\textbf{y}, \textbf{p})=-\sum_{i=1}^n y_i \log p_i.
\end{equation}

\textbf{Dynamic Distill Loss:} LwF \cite{p1} is the first successful case of applying knowledge distillation to CIL. Therefore, we directly incorporate this loss. The specific calculation is as follows:
\begin{equation}
\mathcal{L}_{\text {distill }}=\sum_{j=1}^{\left|y_{t-1}\right|}-S_k\left(f^{t-1}(x)\right) \log S_k(f(x)).
\end{equation}

Due to the varying number of new and old classes in different incremental stages \cite{p51}, we introduce a dynamic balancing parameter $\lambda$ and the overall loss is as follows: :
\begin{equation}
\mathcal{L}=(1-\lambda) \mathcal{L}_{c e}+\lambda \mathcal{L}_{d i s t i l l}\label{eq:example},
\end{equation}
where $\lambda=\left|\mathrm{y}_{\mathrm{t}-1}\right| /\left|\mathrm{y}_{\mathrm{t}}\right|$ represents the proportion of old classes among all classes. It increases as the incremental tasks evolve, indicating that the model focuses more on the old tasks.

\section{Experiments}
\label{sec4}
\subsection{Experimental Setup}
\subsubsection{Datasets}
The experiments were conducted on the CIFAR100 \cite{p52} and ImageNet100 \cite{p48} datasets. The CIFAR100 dataset contains 100 classes, with each class consisting of 600 color images of resolution 32 $\times$ 32 pixels. Each class contains 500 training samples and 100 test samples. ImageNet100 consists of samples from 100 classes, with each sample having a size of 224 $\times$ 224 pixels. Each class contains approximately 1300 training samples and 50 test samples. Our task division follows the two experimental configurations used in DER \cite{p52}: B0, where all classes are evenly distributed across different tasks, and B50, where the first task contains 50 classes (half of the dataset) and the remaining classes are evenly distributed across subsequent tasks. Inc denotes the number of incremental classes.

\subsubsection{Evaluation Metrics}

In the t-th incremental phase, the incremental accuracy $A_{ t }$ denotes the classification accuracy of the current model over all encountered classes up to that point. To facilitate fair comparison across different methods, the average incremental accuracy is computed as the mean accuracy across all incremental phases: $  Avg  = ( 1 / T ) \sum_{ i = 1 } ^ { T } A_{ i }$, where $T$ is the total number of continual learning phases. The $Last$ metric represents the overall accuracy after completing the final task.

\subsubsection{Implementation Details}
We implemented the above method using PyTorch on a NVIDIA RTX 4090 GPU. We use the ViT-B/16 model as the backbone for CLIP. The optimizer is Adam, and each task is trained for 15 epochs. The initial learning rate is 0.001, and it is reduced by a factor of 0.1 at the 4th and 10th epochs. The batch sizes for the ImageNet100 and CIFAR100 datasets are 128 and 100, respectively. For each task, we generate approximately 2,000 samples by sampling from the Gaussian distribution of existing samples to simulate replay samples. The value of $\alpha$ is set to 0.8, and $\beta$ is set to 0.55 for ImageNet100 and 0.65 for CIFAR100. Under the B0 experimental configuration, we set up incremental settings with 5, 10, and 25 tasks. Under the B50 configuration, we set up incremental settings with 5 and 10 tasks. The experimental results of the comparison methods are obtained by running the respective public codes.

\begin{table*}[htbp]
\centering
\caption{Results on the ImageNet100 dataset.}
\label{tab1}
\renewcommand{\arraystretch}{1.15}
\setlength{\tabcolsep}{5pt} %
\begin{tabular}{@{}l
                p{1.2cm}p{1.2cm}|
                p{1.2cm}p{1.2cm}|
                p{1.2cm}p{1.2cm}|
                p{1.2cm}p{1.2cm}|
                p{1.2cm}p{1.2cm}@{}}
\toprule
\multirow{3}{*}{\textbf{Method}} & \multicolumn{6}{c|}{\textbf{ImageNet100 B0}} & \multicolumn{4}{c}{\textbf{ImageNet100 B50}} \\ 
\cmidrule(lr){2-7} \cmidrule(lr){8-11}
 & \multicolumn{2}{c|}{\textbf{Inc5}} & \multicolumn{2}{c|}{\textbf{Inc10}} & \multicolumn{2}{c|}{\textbf{Inc20}} & \multicolumn{2}{c|}{\textbf{Inc5}} & \multicolumn{2}{c}{\textbf{Inc10}} \\ 
\cmidrule(lr){2-3} \cmidrule(lr){4-5} \cmidrule(lr){6-7} \cmidrule(lr){8-9} \cmidrule(lr){10-11}
 & \textit{Avg} & \textit{Last} & \textit{Avg} & \textit{Last} & \textit{Avg} & \textit{Last} & \textit{Avg} & \textit{Last} & \textit{Avg} & \textit{Last} \\ 
\midrule
iCaRL \cite{p45} & 69.00 & 50.90 & 74.10 & 58.50 & 78.10 & 65.20 & 57.30 & 44.40 & 60.70 & 44.70 \\
LUCIR \cite{p46} & 64.70 & 47.80 & 70.50 & 55.30 & 76.00 & 64.00 & 66.90 & 56.80 & 77.20 & 68.20 \\
PODNet \cite{p47} & 66.70 & 48.90 & 72.30 & 72.00 & 78.20 & 66.20 & 79.00 & 70.80 & 80.30 & 73.50 \\ \hline
Cont-CLIP \cite{p15} & 85.74 & 75.40 & 84.98 & 75.40 & 84.03 & 75.40 & 81.35 & 75.40 & 81.09 & 75.40 \\
DualPrompt \cite{p18} & 75.40 & 61.10 & 80.65 & 67.38 & 84.65 & 74.24 & 62.10 & 22.36 & 74.20 & 49.78 \\
L2P++ \cite{p44} & 75.43 & 62.10 & 80.51 & 67.22 & 84.12 & 73.70 & 62.00 & 22.15 & 74.11 & 49.46 \\
ADAM \cite{p48} & 85.78 & 75.72 & 85.84 & 76.40 & 85.85 & 77.08 & 84.90 & 78.58 & 84.60 & 78.58 \\
RARF \cite{p19} & 87.59 & \textbf{79.87} & 87.51 & 80.23 & 86.72 & 80.10 & \textbf{86.53} & \textbf{80.16} & \textbf{86.36} & 80.22 \\ 
CODA \cite{p16} & 51.64 & 24.94 & 64.13 & 34.76 & 69.78 & 43.96 & 57.33 & 19.95 & 65.14 & 28.80 \\
PROOF \cite{p49} & 86.92 & 75.52 & 84.71 & 72.48 & 81.92 & 68.56 & 84.16 & 74.44 & 82.78 & 71.04 \\ \hline
\textbf{CalFuse(Ours)} & \textbf{88.04} & 78.32 & \textbf{88.19} & \textbf{80.32} & \textbf{87.62} & \textbf{81.00} & 85.50 & 79.70 & 85.86 & \textbf{80.40} \\ 
\bottomrule
\end{tabular}
\noindent\parbox{\textwidth}{\raggedright\footnotesize %
}
\end{table*}

\subsection{Comparison with State-Of-The-Art Methods}

Experimental results demonstrate that our method consistently outperforms existing approaches across various class-continual learning settings on large-scale datasets, including CIFAR100 and ImageNet100. The compared methods are broadly divided into two categories: traditional CCL methods and non-traditional CCL methods. Specifically, the results of the Continual-CLIP method correspond to the zero-shot performance of the CLIP model. To ensure a fair and rigorous comparison, all non-traditional CCL baselines are implemented using the same CLIP pretrained weights, maintaining consistency in large-scale multimodal feature representations.
\begin{table*}[htbp]
\centering
\caption{Results on the CIFAR100 dataset.}
\label{tab2}
\renewcommand{\arraystretch}{1.15} %
\setlength{\tabcolsep}{5pt}        %
\begin{tabular}{@{}l
                p{1.2cm}p{1.2cm}|
                p{1.2cm}p{1.2cm}|
                p{1.2cm}p{1.2cm}|
                p{1.2cm}p{1.2cm}|
                p{1.2cm}p{1.2cm}@{}}
\toprule
\multirow{3}{*}{\textbf{Method}} & \multicolumn{6}{c|}{\textbf{CIFAR100 B0}} & \multicolumn{4}{c}{\textbf{CIFAR100 B50}} \\ 
\cmidrule(lr){2-7} \cmidrule(lr){8-11}
 & \multicolumn{2}{c|}{\textbf{Inc5}} & \multicolumn{2}{c|}{\textbf{Inc10}} & \multicolumn{2}{c|}{\textbf{Inc20}} & \multicolumn{2}{c|}{\textbf{Inc5}} & \multicolumn{2}{c}{\textbf{Inc10}} \\ 
\cmidrule(lr){2-3} \cmidrule(lr){4-5} \cmidrule(lr){6-7} \cmidrule(lr){8-9} \cmidrule(lr){10-11}
 & \textit{Avg} & \textit{Last} & \textit{Avg} & \textit{Last} & \textit{Avg} & \textit{Last} & \textit{Avg} & \textit{Last} & \textit{Avg} & \textit{Last} \\ 
\midrule
iCaRL \cite{p45} & 61.20 & 44.90 & 65.30 & 50.90 & 71.10 & 59.60 & 58.60 & 49.50 & 65.10 & 56.00 \\
LUCIR \cite{p46} & 58.20 & 41.10 & 58.70 & 42.90 & 62.80 & 46.90 & 59.90 & 48.20 & 64.30 & 52.70 \\
PODNet \cite{p47} & 54.00 & 35.80 & 58.00 & 40.70 & 66.70 & 51.50 & 64.00 & 51.70 & 67.30 & 55.90 \\ \hline
DualPrompt \cite{p18} & 79.74 & 69.91 & 81.45 & 72.51 & 85.19 & 77.47 & 58.55 & 15.26 & 72.00 & 45.05 \\
Cont-CLIP \cite{p15} & 75.93 & 66.68 & 75.00 & 66.68 & 74.01 & 66.68 & 70.79 & 66.68 & 70.77 & 66.68 \\
L2P++ \cite{p44} & 79.18 & 68.67 & 81.90 & 73.08 & 84.39 & 77.37 & 58.57 & 18.04 & 76.51 & 48.52 \\
ADAM \cite{p48} & 70.18 & 58.12 & 80.53 & 65.50 & 77.28 & 67.89 & 83.38 & 76.94 & 83.21 & 76.94 \\
RARF \cite{p19} & 86.61 & 78.68 & 86.01 & 79.27 & \textbf{85.65} & 79.82 & \textbf{83.19} & \textbf{79.45} & 83.06 & \textbf{79.85} \\
PROOF \cite{p49} & 85.12 & 76.13 & 84.88 & 76.29 & 84.11 & 76.86 & 83.22 & 76.25 & 83.17 & 76.50 \\
CODA \cite{p16} & 69.78 & 41.98 & 76.98 & 62.25 & 78.65 & 65.29 & 58.45 & 15.99 & 67.88 & 28.77 \\ \hline
\textbf{CalFuse(Ours)} & \textbf{86.70} & \textbf{78.84} & \textbf{86.06} & \textbf{79.35} & 85.45 & \textbf{79.87} & 83.04 & 79.38 & \textbf{83.26} & 79.58 \\ 
\bottomrule
\end{tabular}
\end{table*}

\subsubsection{Results on ImageNet100}
Table~\ref{tab1} presents the performance of our proposed method on the ImageNet100 dataset under five distinct continual learning settings. It is evident from the results that our approach, which deviates from conventional CCL paradigms, significantly outperforms traditional methods. The observed performance gap is substantial and consistent across all settings, highlighting the effectiveness of integrating CLIP into the CCL framework. This performance boost is largely attributed to the incorporation of text features, which enhance the semantic alignment between visual and textual modalities. These results underscore the promising potential of CLIP in continual learning scenarios, where maintaining performance across sequential tasks is notoriously challenging.
Compared to the zero-shot performance of CLIP, our method achieves average incremental accuracy gains of 2.3\%, 3.21\%, 3.59\%, 4.15\%, and 4.77\% on the B0Inc5, Inc10, Inc20, and B50Inc5, Inc10 incremental task settings, respectively. It suggests that our framework does not rely solely on the strong generalization capabilities of pre-trained CLIP representations. Instead, the gains reflect the effectiveness of our novel FC and PF strategies, which are specifically designed to address the challenges of catastrophic forgetting and representation shift in CCL.
Compared to DualPrompt \cite{p18} and CODA \cite{p16}, our method consistently achieves superior results. While DualPrompt and CODA focus primarily on optimizing the textual prompt space, our findings suggest that improvements in prompt engineering alone are insufficient for robust CCL performance. Our approach leverages both textual and visual modalities by enhancing visual feature utilization, demonstrating that a more holistic integration of multimodal representations is critical.
Under the B0 experimental setting, our method achieve average accuracy of 88.04\%, 88.19\%, 87.62\%, exceeding the next-best methods by margins of at least 1.12\%, 2.35\%, and 1.77\%, respectively. These consistent gains further validate the robustness and superiority of our proposed techniques. Additionally, when evaluating the performance based on the final task accuracy—a critical indicator of a model’s resistance to forgetting—our method maintains a performance level around 80\%. In contrast, other methods exhibit significantly lower final accuracies, indicating greater susceptibility to catastrophic forgetting.
These findings confirm that our method offers a well-balanced and effective solution for CCL. The integration of parameter decomposition, feature fusion, and dynamic distillation not only mitigates the drawbacks of previous approaches but also introduces a resilient framework that adapts well to evolving class distributions.  

\subsubsection{Results on CIFAR100}

Table~\ref{tab2} presents the performance evaluation of our proposed method on the CIFAR-100 dataset under five distinct continual learning scenarios. As shown, non-traditional CCL approach significantly outperforms existing traditional methods, exhibiting an even more pronounced performance gap compared to the results observed on the ImageNet100 dataset. This substantial improvement underscores the generalizability and robustness of our approach across different datasets, including those with smaller resolution and less semantic complexity like CIFAR-100. 
Compared to the zero-shot performance of CLIP, our method achieves average incremental accuracy gains of 10.77\%, 11.06\%, 11.44\%, 12.25\%, and 12.49\% on the B0Inc5, Inc10, Inc20, and B50Inc5, Inc10 incremental task settings, respectively. These gains are approximately two to three times greater than those observed on ImageNet100, further confirming the effectiveness of the proposed FC and PF mechanisms. 
Under the B0 experimental setting, our method achieve average accuracy of 86.70\%, 86.06\%, and 85.45\% across different incremental task settings. These results are consistently higher than those achieved by competing methods, providing strong evidence for the robustness and superiority of our approach. Furthermore, the last metric, which indicates the final task performance and serves as a key measure of a model's resilience to catastrophic forgetting, remains consistently above 79.30\% across all experimental settings. In contrast, other methods are nearly unattainable to reach this level, often experiencing significant performance degradation over time. This consistent and high final performance highlights the effectiveness of our method in mitigating the forgetting problem—a central challenge in CCL. Overall, the outstanding performance of our method on CIFAR-100, demonstrates that our approach not only scales well across datasets but also provides a powerful and general framework for CCL.

\subsection{Ablation Studies}
\subsubsection{Module Ablation}

\begin{table*}%
\centering
\caption{Results of module ablation on the ImageNet100 B0 Inc10.}
\resizebox{\textwidth}{!}{%
\begin{tabular}{cccccccccccccccc}
\toprule
 & \textbf{FC} & \textbf{PF} & \textbf{Distill} & \textbf{$A_{ 1 }$} & \textbf{$A_{ 2 }$} & \textbf{$A_{ 3 }$} & \textbf{$A_{ 4 }$} & \textbf{$A_{ 5 }$} & \textbf{$A_{ 6 }$} & \textbf{$A_{ 7 }$} & \textbf{$A_{ 8 }$} & \textbf{$A_{ 9 }$} & \textbf{$A_{ 10 }$ (\textit{Last})} & \textbf{\textit{Avg}}  \\
\midrule
1 & \xmark & \xmark & \xmark & 97.20 & 95.20 & 91.93 & 88.75 & 83.56 & 81.80  & 81.00   & 78.89 & 76.10  & \textbf{75.40}  & \textbf{84.98}  \\
2 & \cmark & \xmark & \xmark & 96.6 & 97.00 & 92.43 & 90.80  & 83.64 & 84.13 & 82.27 & 80.97 & 76.56 & \textbf{73.78} & \textbf{85.82}  \\
3 & \cmark & \cmark & \xmark & 96.80 & 96.60 & 93.87 & 91.70  & 87.20  & 85.52 & 85.03 & 84.60  & 80.27 & \textbf{80.02} & \textbf{88.16}  \\
4 & \cmark & \cmark & \cmark & 96.80 & 96.10 & 94.27 & 91.50  & 87.16 & 85.43 & 85.09 & 84.82 & 80.42 & \textbf{80.32} & \textbf{88.19}  \\
\bottomrule
\end{tabular}%
}
\label{tab3}
\end{table*}

Table~\ref{tab3} presents the results of our ablation study, which evaluates the individual contributions of each module in our proposed method. Overall, our method demonstrates stable performance throughout all incremental stages. Meanwhile, the table shows the role of FC, PF, and distillation in enhancing model performance and mitigating catastrophic forgetting. The inclusion of the FC module leads to a 0.84\% increase in average incremental accuracy, demonstrating that preserving the integrity of the original CLIP features is highly beneficial. FC prevents the distortion of pre-trained visual embeddings that often results from direct fine-tuning, thereby retaining the strong generalization capacity of CLIP. However, we observe a 1.62\% drop in the last metric when FC is used alone, indicating that while it facilitates the acquisition of new knowledge, it may introduce a trade-off in terms of long-term memory retention. This suggests that FC enhances short-term task performance but slightly exacerbates forgetting over time if used in isolation.
PF shows a more substantial contribution. It uses QR decomposition to retain old knowledge, improves accuracy by 2.34\% and the last metric by 6.24\%. These results clearly indicate that PF not only supports effective learning of new tasks but also plays a critical role in preserving previously learned knowledge. This module leverages the orthogonality principle in QR decomposition to facilitate better knowledge retention across tasks. The improvement in the last metric suggests that decomposing and fusing the adapter’s parameter matrix is a highly effective strategy, achieving a delicate balance between plasticity (learning new tasks) and stability (retaining old knowledge). The dynamic distillation module plays a more prominent role in mitigating the forgetting problem. By adaptively distilling knowledge from earlier task models and dynamically updating the student model, this module reduces the drift of representations and promotes consistency across learning stages. Its contribution becomes especially evident in the sustained high values of the last metric when combined with other modules, underscoring its vital role in preventing catastrophic forgetting. 
Overall, the ablation study validates the effectiveness of each individual module in our framework. While FC aids in preserving feature quality and learning new information, PF ensures that old knowledge is not overwritten, and dynamic distillation offers a strong regularization mechanism for continual learning. The synergy among these components forms the foundation of our robust and adaptive CCL framework.

\subsubsection{Effectiveness of Loss Function}
In Eq.\ref{eq:example}, our loss function is composed of two key components: the standard cross-entropy loss, which facilitates the learning of new class information, and a dynamic distillation loss, which serves to preserve knowledge from previously learned tasks. These two components are weighted by a scalar parameter $\lambda$, which plays a crucial role in balancing the influence of old and new class information during training. As evidenced by the ablation results in Table\ref{tab3}, this design significantly impacts the model’s ability to mitigate forgetting and retain prior knowledge. 
When only the cross-entropy loss is used—without incorporating any distillation mechanism—the final task accuracy (last) is 80.02\%. This baseline reflects the typical performance of a model focused solely on learning new information without explicit mechanisms to retain prior knowledge. However, when we introduce the distillation loss alongside the cross-entropy loss, and dynamically modulate their relative contributions using the parameter $\lambda$, the last metric improves to 80.32\%. Although the numerical gain may appear modest, it demonstrates that even a small degree of incremental knowledge transfer can contribute meaningfully to long-term retention in CCL.
The parameter $\lambda$ is not static, rather, it is dynamically computed based on the ratio of old classes to the total number of classes seen so far in the continual learning process. This adaptive strategy ensures that as the task progresses and the balance between old and new classes changes, the loss function remains sensitive to the shifting importance of knowledge retention versus acquisition. In contrast, when $\lambda$ is fixed across all learning stages, the model’s accuracy decreases. This drop illustrates the limitation of static weighting schemes in dynamic learning environments.
The results clearly indicate that the dynamic design of $\lambda$ contributes to better performance by providing an adaptive mechanism that reflects the class distribution at each stage. As the number of classes increases, and the potential for forgetting previously learned classes grows, the dynamic $\lambda$ ensures that the distillation loss is appropriately weighted to preserve old knowledge. Conversely, in early stages with fewer old classes, the model is allowed greater flexibility to focus on new knowledge acquisition. This adaptability allows our loss formulation to strike an effective balance between stability and plasticity, which is essential in continual learning settings.
Overall, the incorporation of a dynamically adjusted $\lambda$ improves the effectiveness of the distillation process. This design choice further reinforces the robustness and adaptability of our proposed framework in addressing the challenges of catastrophic forgetting.

\begin{figure*}[t] %
\centering
\setlength{\tabcolsep}{1pt} %
\renewcommand{\arraystretch}{1.0}

\begin{tabular}{ccccc}
\includegraphics[width=0.19\textwidth]{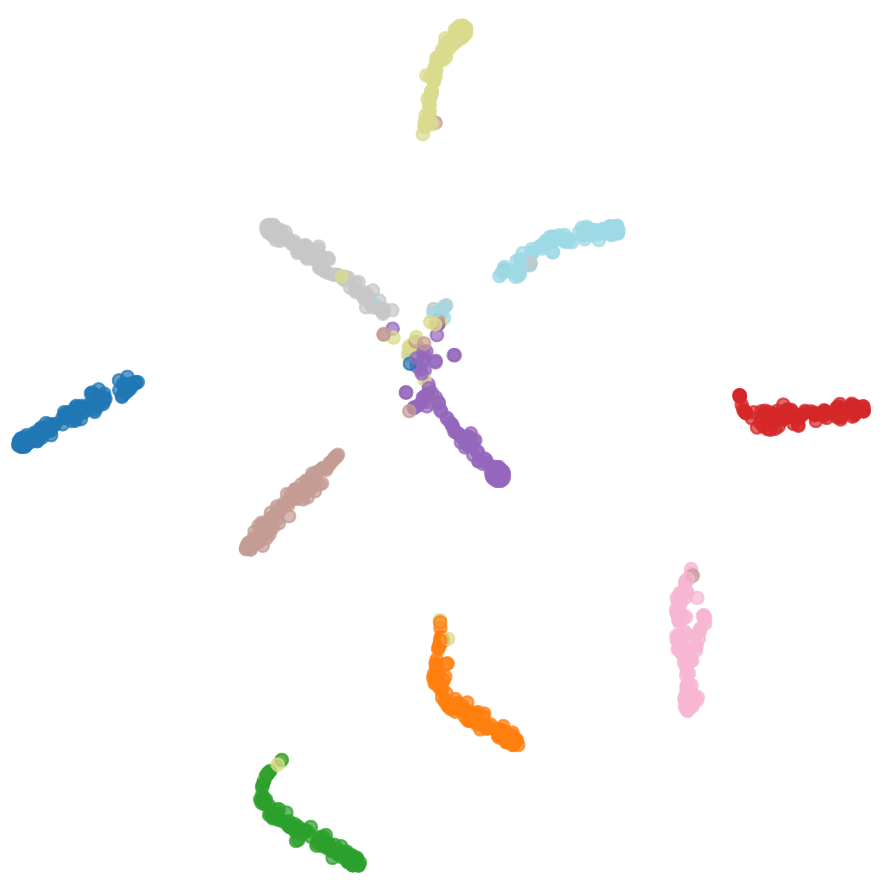} &
\includegraphics[width=0.19\textwidth]{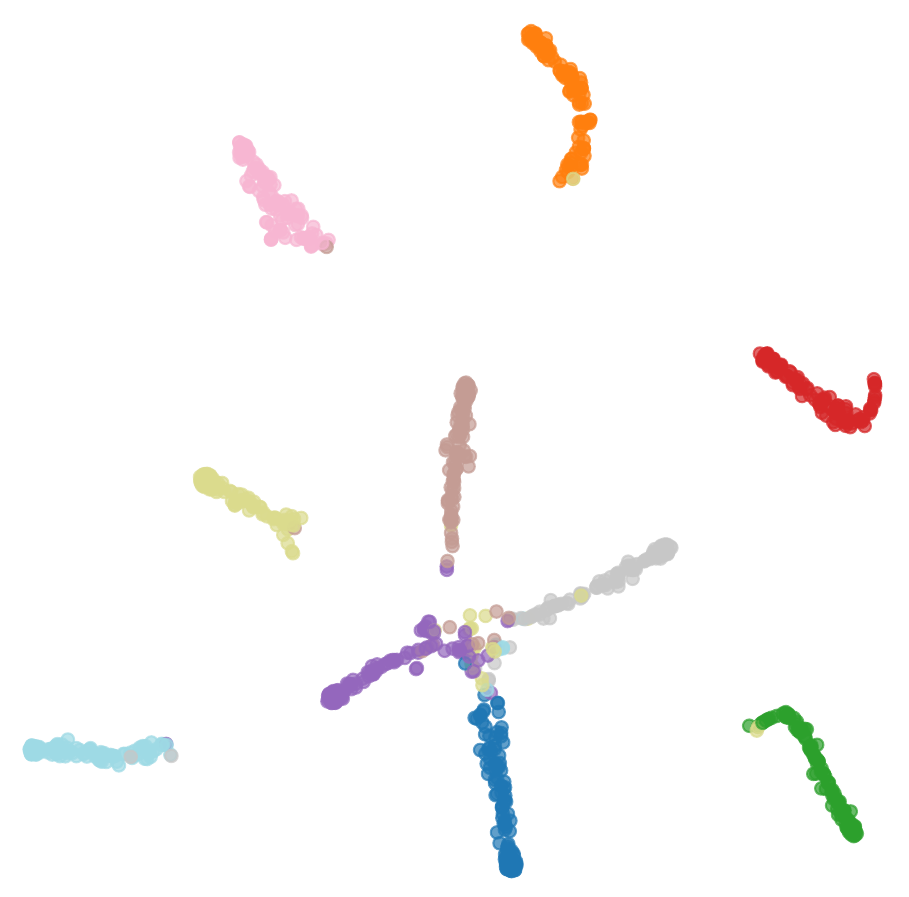} &
\includegraphics[width=0.19\textwidth]{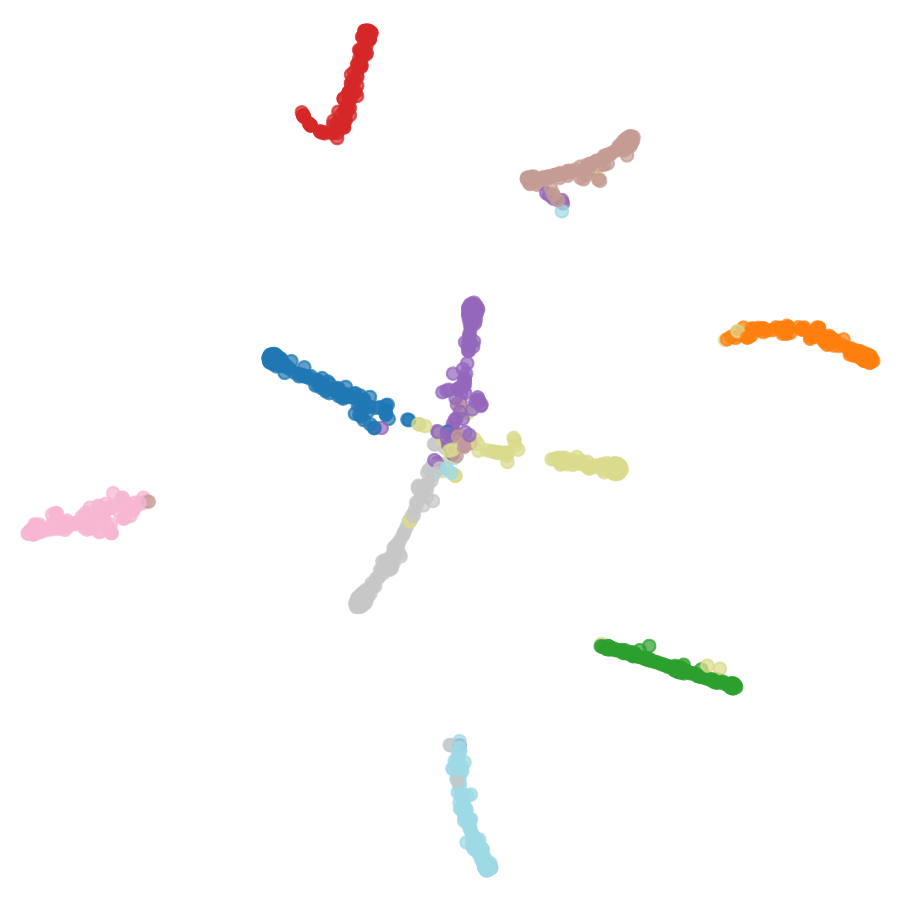} &
\includegraphics[width=0.19\textwidth]{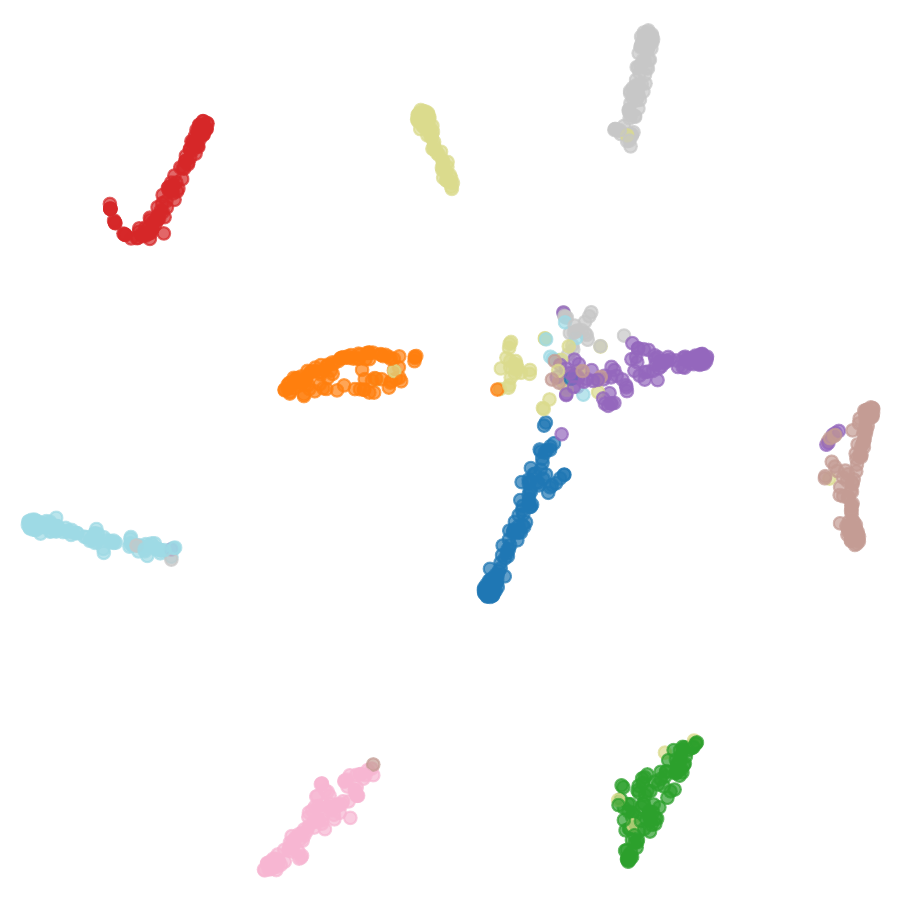} &
\includegraphics[width=0.19\textwidth]{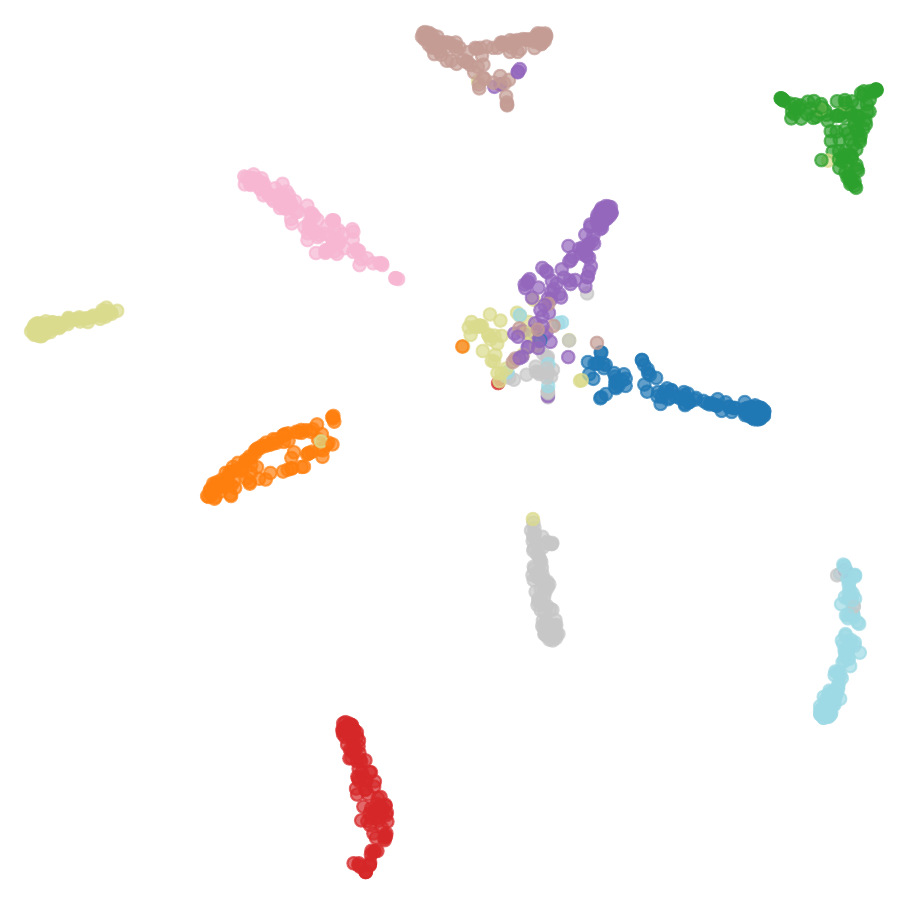} \\[-2pt]
Task 1 & Task 2 & Task 3 & Task 4 & Task 5
\end{tabular}

\vspace{5pt} %

\begin{tabular}{ccccc}
\includegraphics[width=0.19\textwidth]{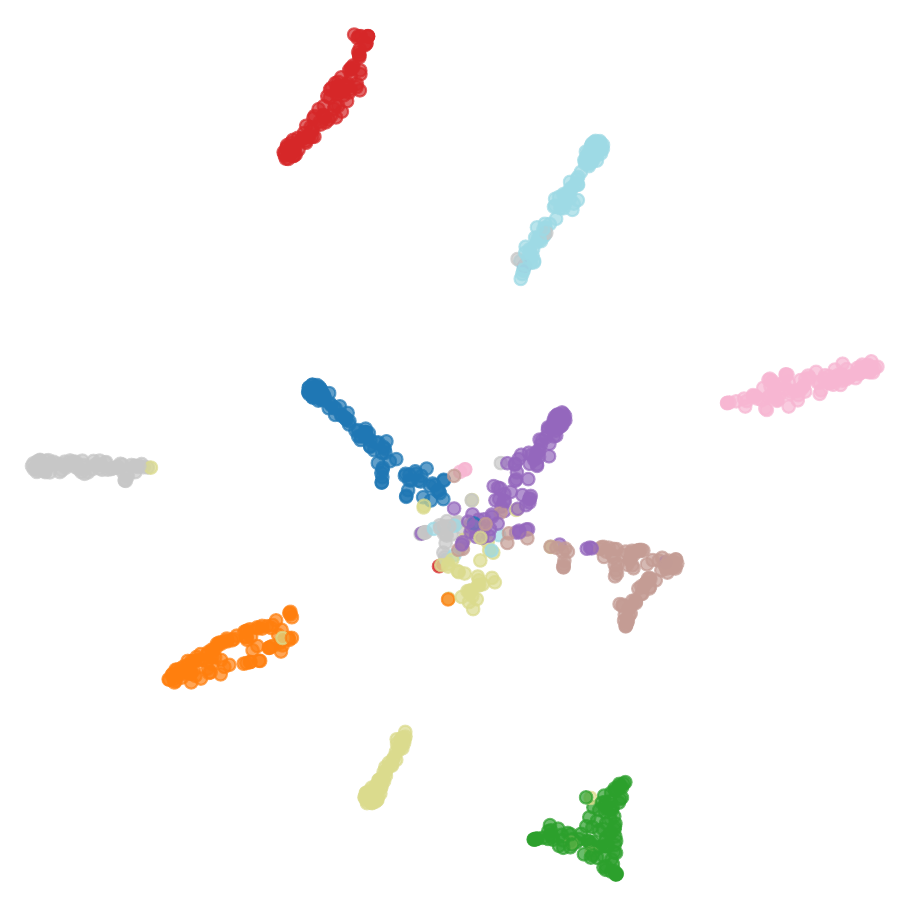} &
\includegraphics[width=0.19\textwidth]{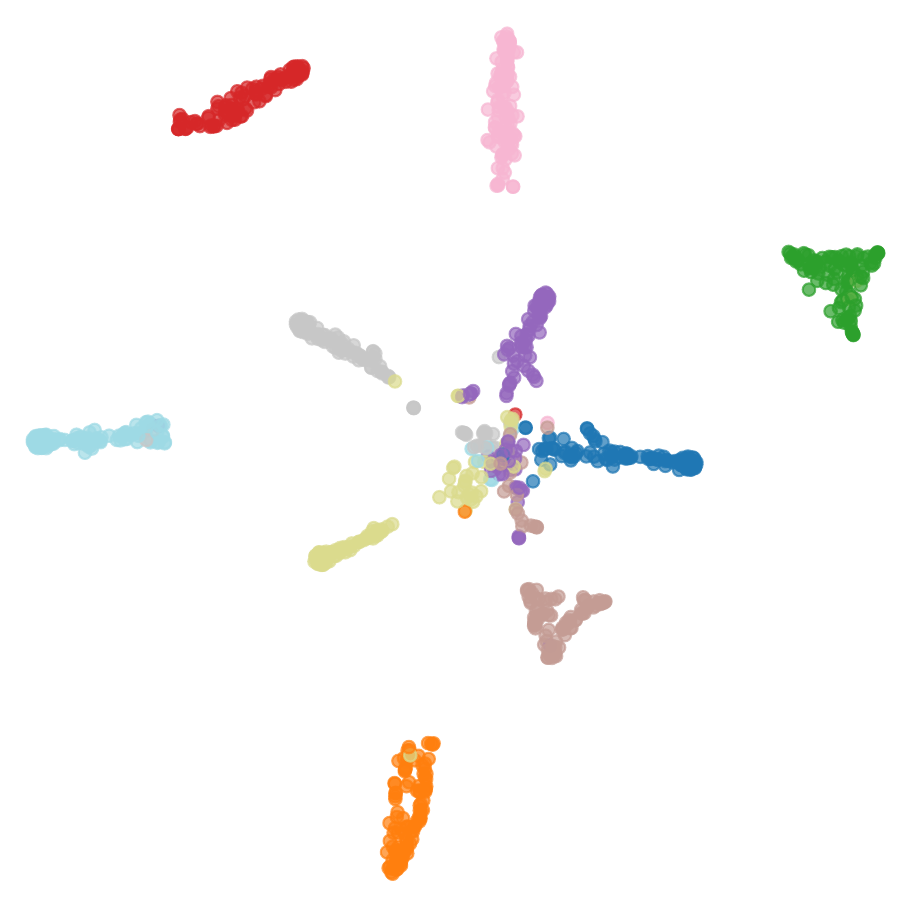} &
\includegraphics[width=0.19\textwidth]{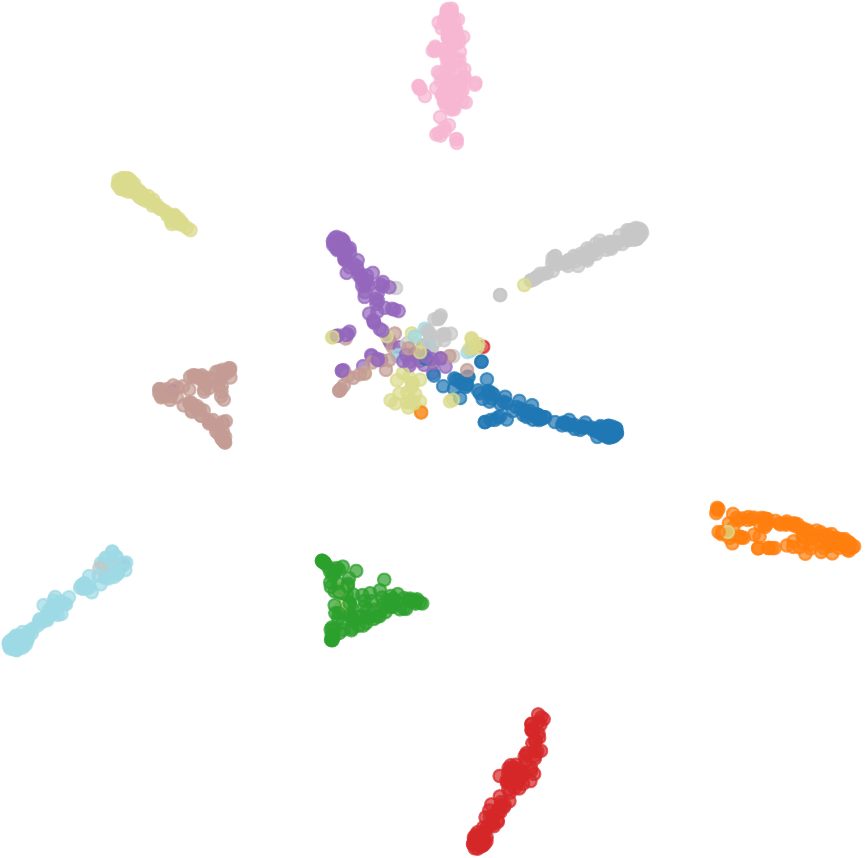} &
\includegraphics[width=0.19\textwidth]{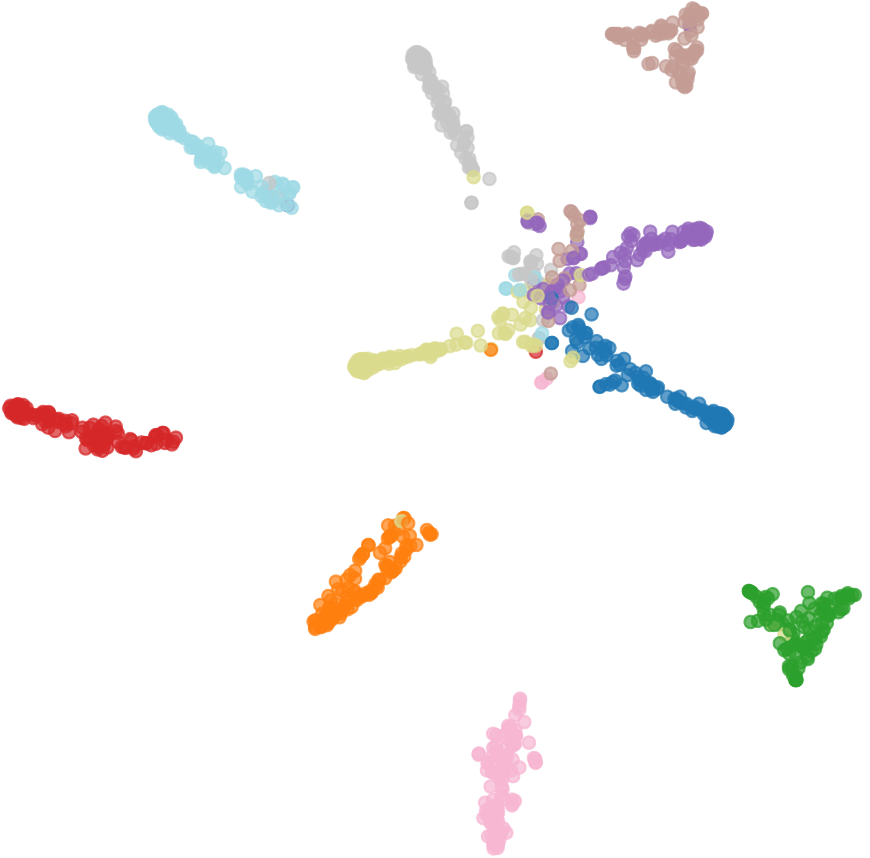} &
\includegraphics[width=0.19\textwidth]{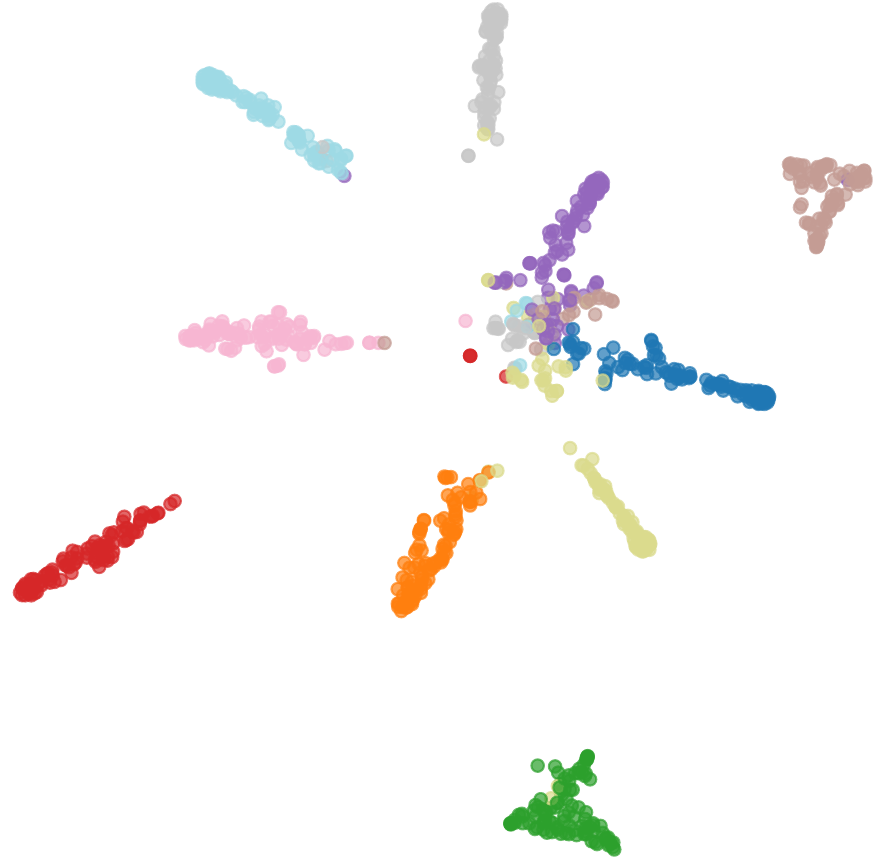} \\[-2pt]
Task 6 & Task 7 & Task 8 & Task 9 & Task 10
\end{tabular}

\caption{T-SNE visualization results of CLIP with ImageNet100 across ten continual learning tasks. The classes shown in each figure are from the ten classes of the first task. After learning each new task, the feature space quality of the model on the 10 classes of the first task is visualized.
}
\label{figa}
\end{figure*}

\begin{figure*}[t] %
\centering
\setlength{\tabcolsep}{1pt} %
\renewcommand{\arraystretch}{1.0}

\begin{tabular}{ccccc}
\includegraphics[width=0.19\textwidth]{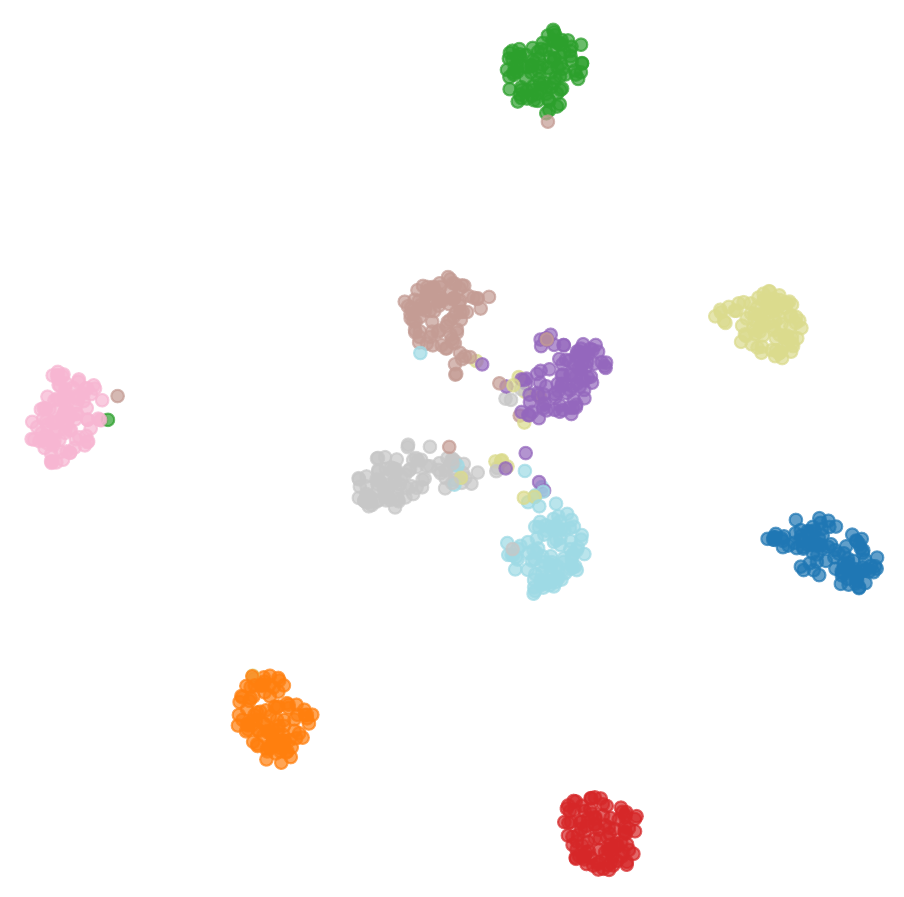} &
\includegraphics[width=0.19\textwidth]{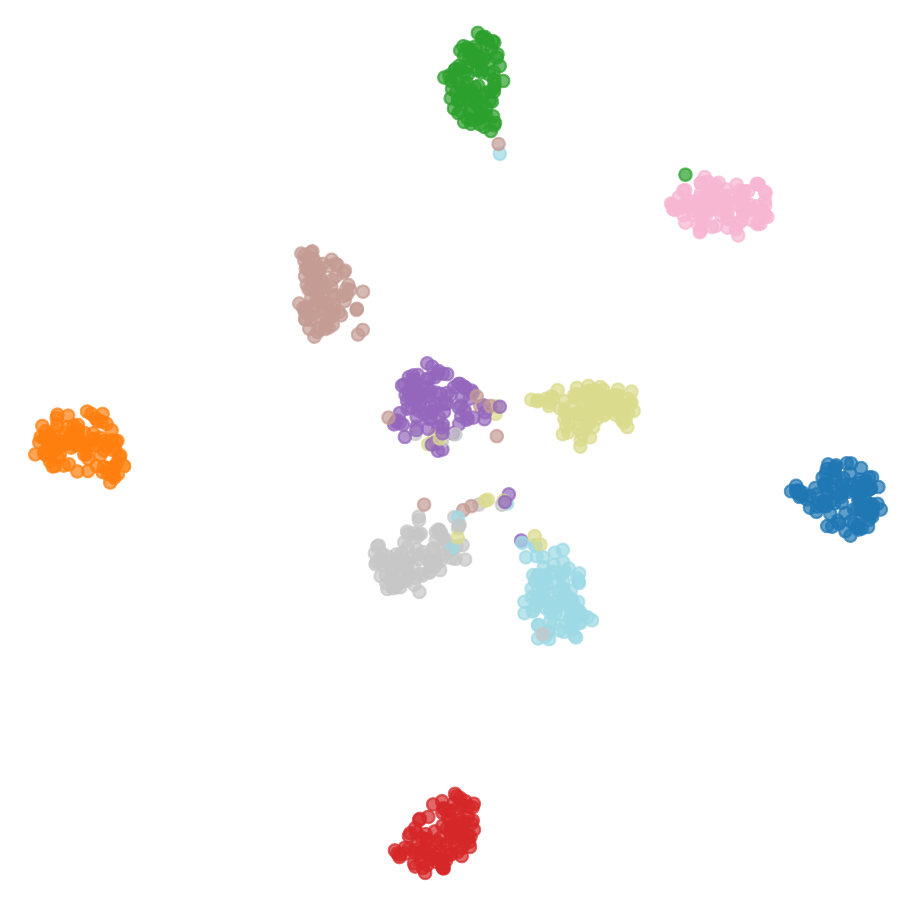} &
\includegraphics[width=0.19\textwidth]{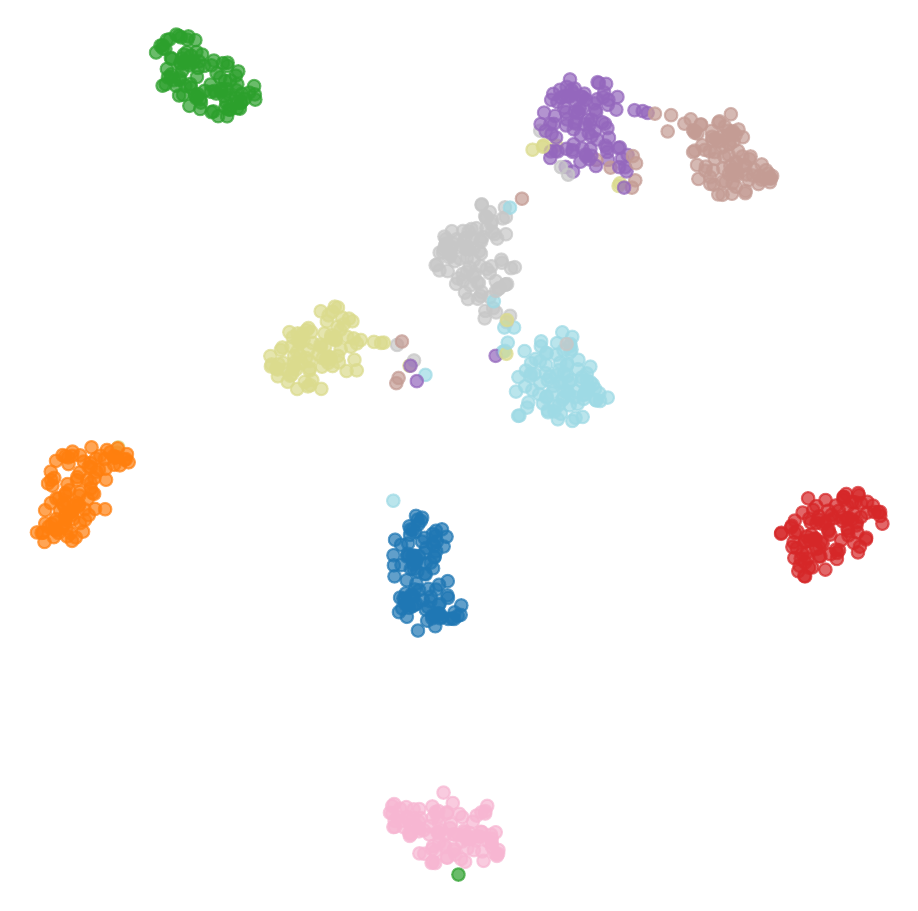} &
\includegraphics[width=0.19\textwidth]{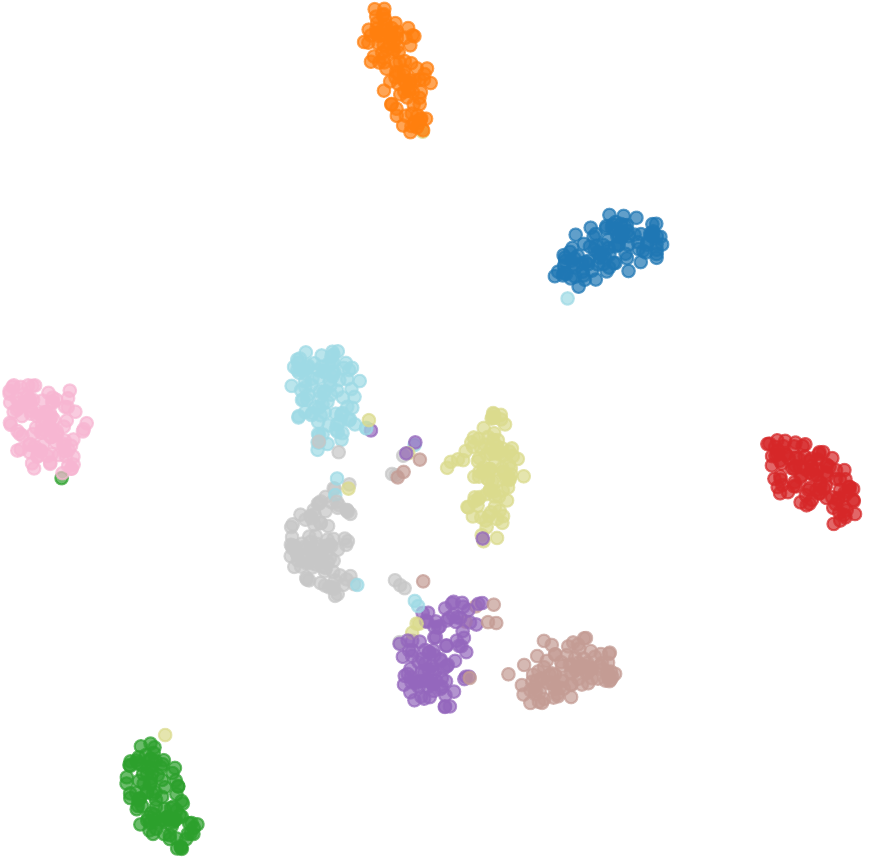} &
\includegraphics[width=0.19\textwidth]{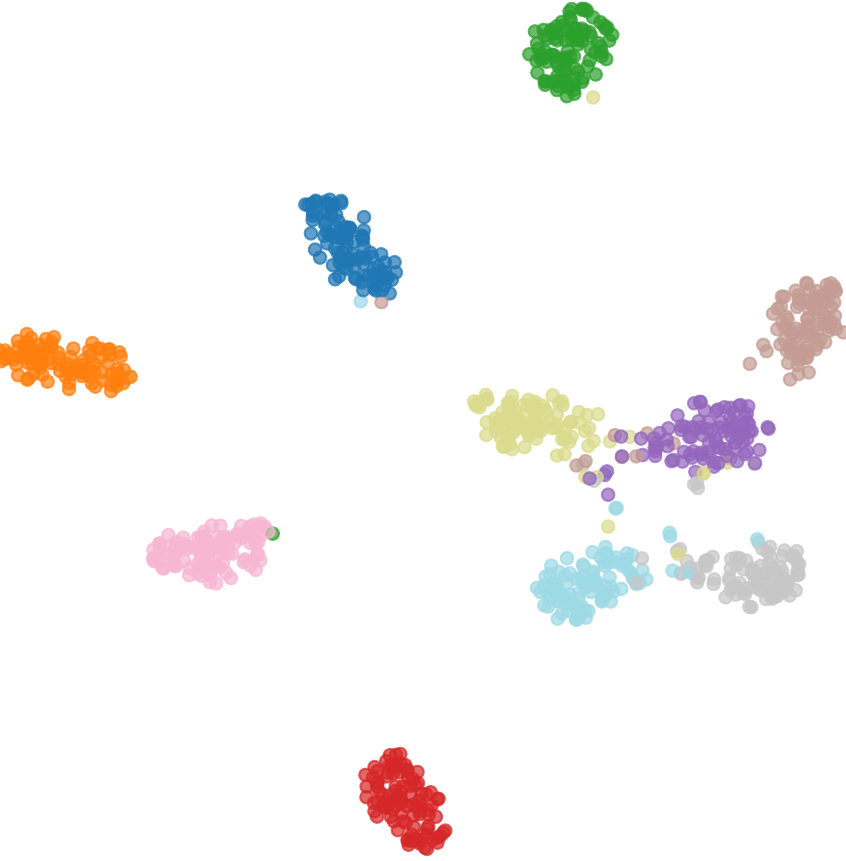} \\[-2pt]
Task 1 & Task 2 & Task 3 & Task 4 & Task 5
\end{tabular}

\vspace{5pt} %

\begin{tabular}{ccccc}
\includegraphics[width=0.19\textwidth]{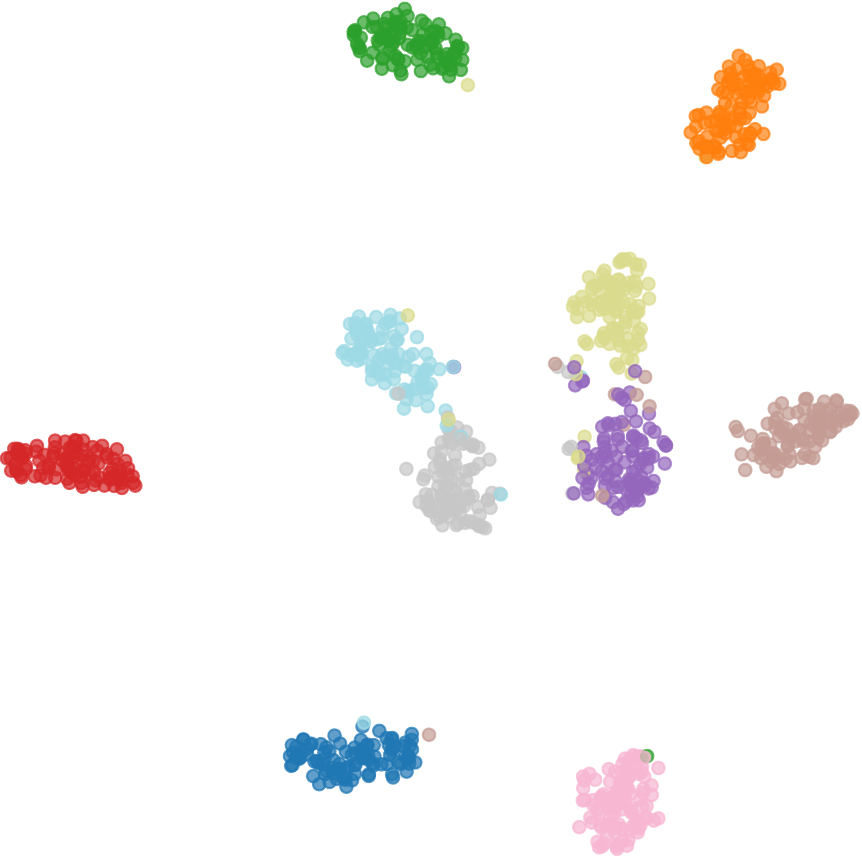} &
\includegraphics[width=0.19\textwidth]{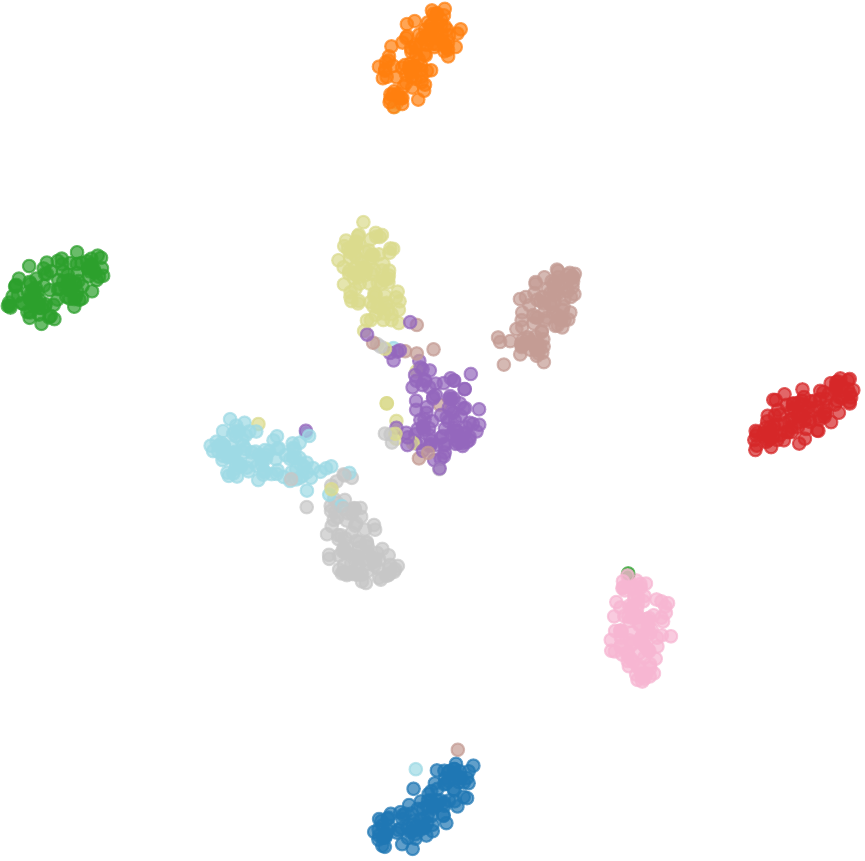} &
\includegraphics[width=0.19\textwidth]{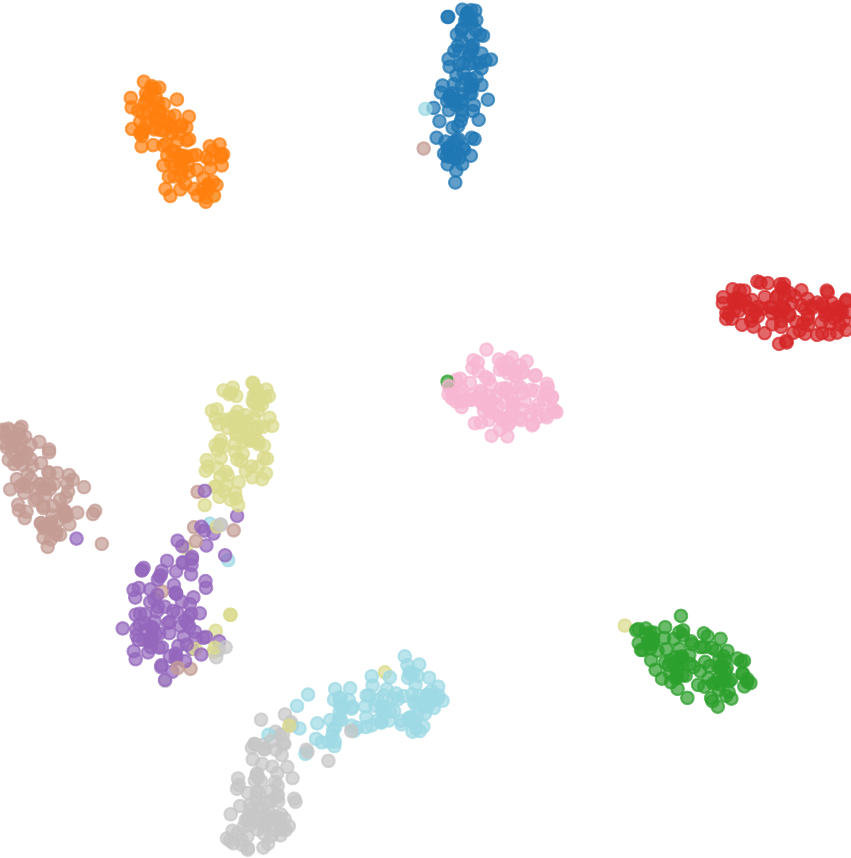} &
\includegraphics[width=0.19\textwidth]{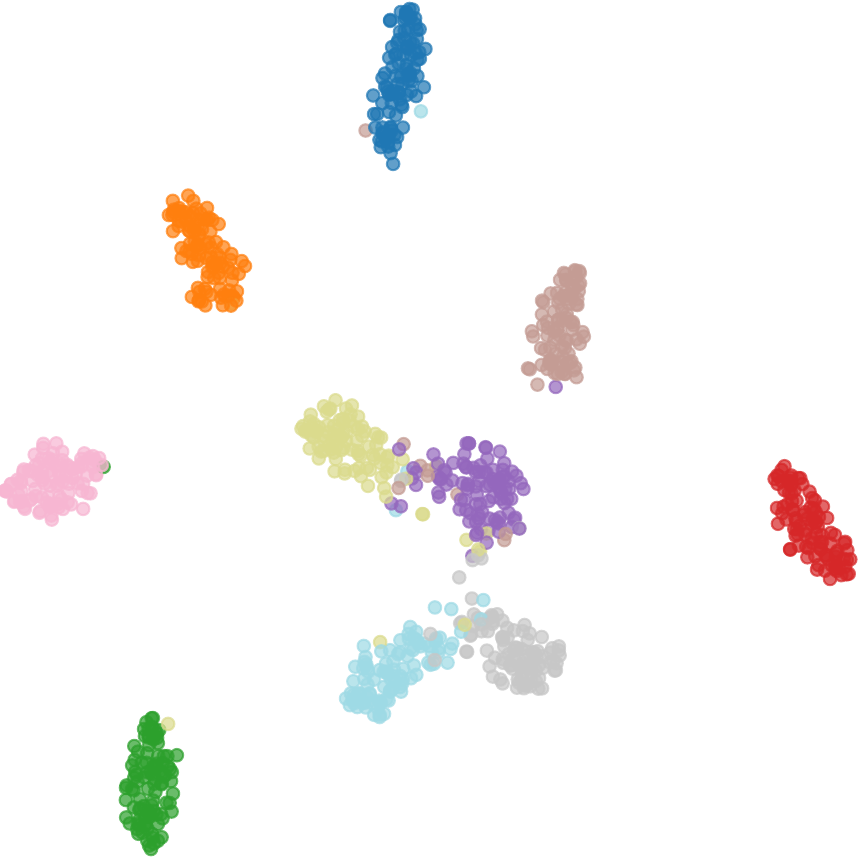} &
\includegraphics[width=0.19\textwidth]{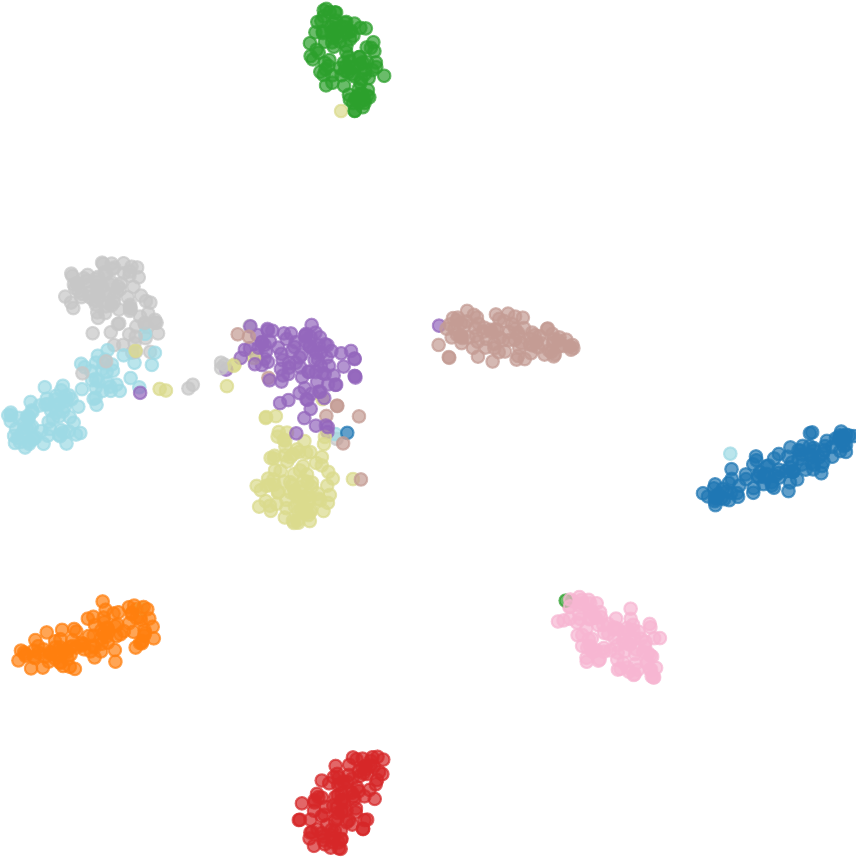} \\[-2pt]
Task 6 & Task 7 & Task 8 & Task 9 & Task 10
\end{tabular}

\caption{T-SNE visualization results of CalFuse with ImageNet100 across ten continual learning tasks.}
\label{figb}
\end{figure*}

\begin{figure}%
  \centering
  \includegraphics[width=0.49\textwidth]{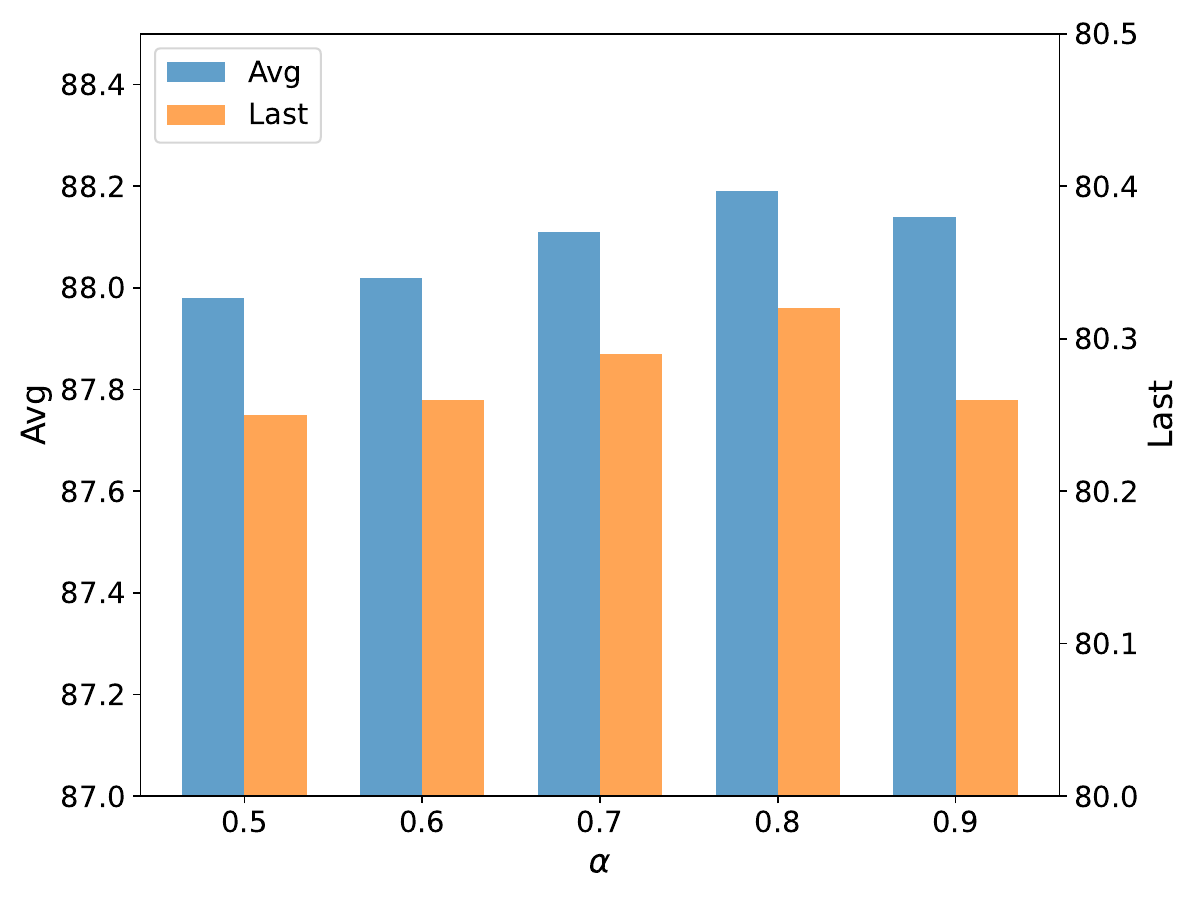}
  \hfill
  \includegraphics[width=0.49\textwidth]{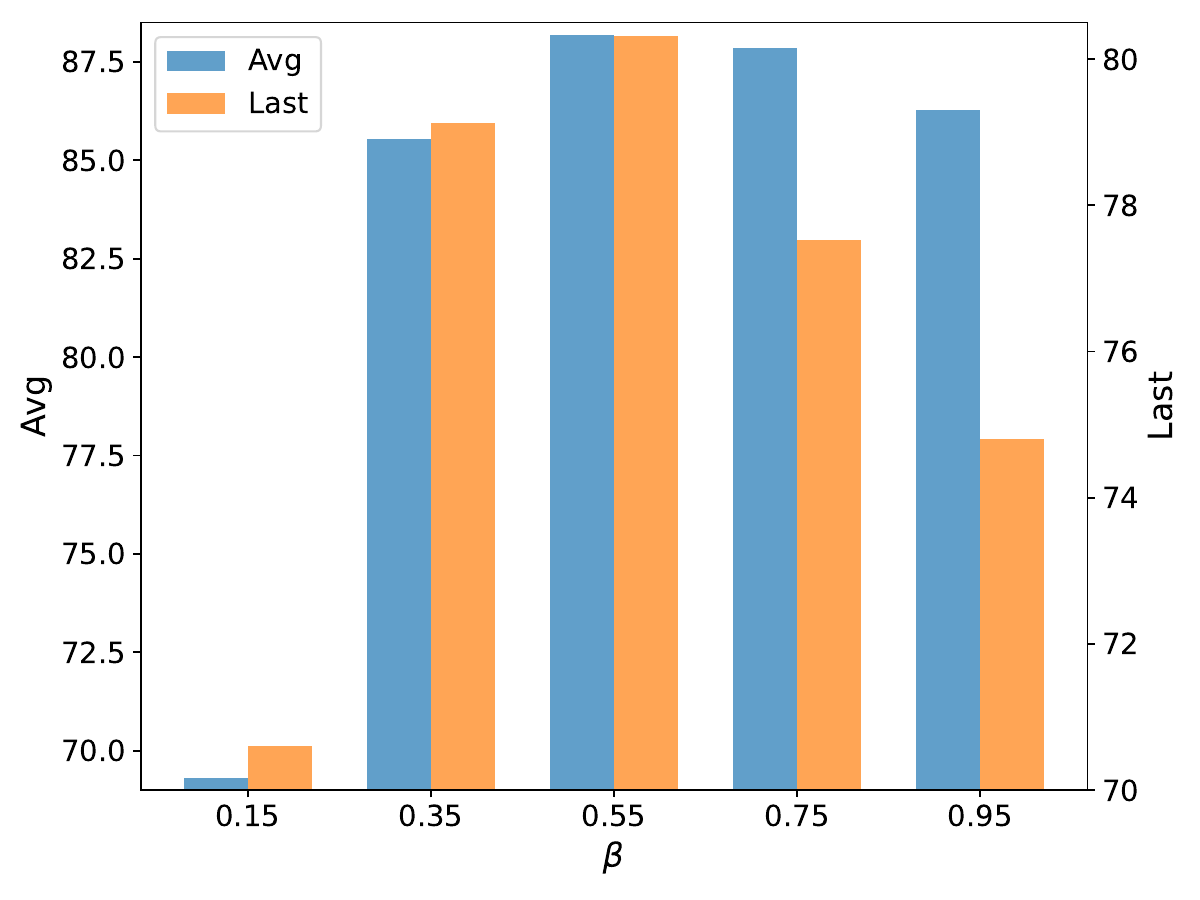}
  \caption{Results of performance sensitivity analysis on (a) $\alpha$ and (b) $\beta$ with ImageNet100 B0 Inc10 configuration.}
  \label{fig4}
\end{figure}

\subsubsection{Visualization results of T-SNE}
Fig.~\ref{figa} and Fig.~\ref{figb} present the t-SNE visualizations of CLIP and CalFuse on the ImageNet100 dataset, illustrating the evolution of the feature space across ten continual learning tasks. Both visualizations show how the representations of the first-task classes change as the model learns new tasks. Overall, CLIP exhibits noticeable feature drift and inter-class overlap during incremental learning, whereas CalFuse demonstrates superior feature preservation and inter-class separability. Specifically, in the early stage (Task 1), CLIP forms distinguishable yet loosely scattered clusters. However, as the learning process continues, the feature space of CLIP gradually deteriorates — the clusters begin to collapse and overlap, and the class boundaries become increasingly blurred. By later stages (Tasks 6–10), multiple class clusters merge or collapse into overlapping regions, indicating a severe degradation of the original feature structure. This phenomenon reflects the catastrophic forgetting problem inherent in CLIP, where the model progressively loses its ability to distinguish features from previously learned tasks, resulting in dispersed intra-class distributions and reduced inter-class discrimination.
In contrast, CalFuse exhibits remarkable stability and robustness in its learned representations under the same experimental setting. From Task 1 to Task 10, CalFuse maintains compact intra-class clusters and well-separated inter-class boundaries, with minimal drift or overlap. The clusters remain consistent and distinct throughout the continual learning process, suggesting that CalFuse effectively preserves old knowledge while integrating new information. This performance gain primarily stems from the proposed stacked parameter fusion and adaptive weighting mechanisms, which jointly mitigate feature interference and drift, thereby enhancing knowledge consistency and feature discriminability across tasks. From an overall perspective, CalFuse produces a more structured and well-organized embedding space with clear inter-class margins, highlighting its stronger discriminative capability and robustness. In contrast, CLIP’s embeddings appear irregular and more prone to interference between classes, suggesting weaker generalization stability under task transitions.
In summary, the t-SNE visualizations clearly validate the superiority of CalFuse in continual learning scenarios. Compared with CLIP, CalFuse achieves significantly higher intra-class compactness and inter-class separability while maintaining a stable and coherent feature space as new tasks are introduced. These results confirm that the integration of stacked parameter fusion and adaptive weighting is both effective and necessary for achieving high-quality feature representation and strong anti-forgetting performance.

\subsubsection{Sensitivity Analysis on $\alpha$ and $\beta$}
In our framework, preserving the original CLIP features is a key design consideration, and we introduce the parameter $\alpha$ to control the degree of feature preservation. Specifically, $\alpha$ serves as a balancing coefficient that governs the trade-off between the original frozen CLIP features and the newly enhanced features derived from task-specific adaptations. As illustrated in Fig.~\ref{fig4}(a), variations in the value of $\alpha$ yield noticeable changes in model accuracy, highlighting the sensitivity of the model’s performance to this balance. This observation underscores the importance of carefully calibrating how much of the pre-trained knowledge is retained versus how much is adapted to new tasks. When $\alpha$ is too small, the model may underutilize the rich semantic representations encoded in CLIP, leading to suboptimal generalization. Conversely, when $\alpha$ is too large, the model may become overly reliant on static features, limiting its ability to adapt to new class distributions in the continual learning setting. Thus, selecting an appropriate value for $\alpha$ is essential to achieving a balanced integration of pre-trained and task-specific knowledge.
In contrast to $\alpha$, the parameter $\beta$, which controls the influence of the decomposed and fused adapter parameters, has an even more direct and pronounced effect on model performance. As depicted in Fig.~\ref{fig4}(b), improper tuning of $\beta$ can significantly degrade accuracy, revealing the sensitivity of the model to the fusion mechanism applied to the adapter’s parameters. This further emphasizes the critical role of parameter decomposition and fusion in our approach. The adapter module, which operates as a lightweight, trainable bridge between frozen CLIP representations and the downstream incremental tasks, relies heavily on the proper scaling and integration of its components. $\beta$ modulates the contribution of these components, and an ill-suited value can disrupt the representational harmony between old and new knowledge, ultimately hindering the model’s capacity to learn incrementally without forgetting.
The experiments shown in Fig.~\ref{fig4} collectively reveal that both $\alpha$ and $\beta$ are vital hyperparameters, but they influence different aspects of the learning process. While $\alpha$ governs the trade-off between stability (retaining CLIP’s pre-trained knowledge) and plasticity (learning task-specific enhancements), $\beta$ directly affects the internal representational learning within the adapter. Optimal performance is achieved only when both parameters are carefully tuned, enabling a synergistic interaction between feature preservation and adaptable parameterization.
\section{Discussion}
\subsection{Limitation}
Although the proposed approach demonstrates promising performance in multimodal continual learning, several limitations remain that warrant further investigation.
First, the current framework relies on manually configured hyperparameters, including learning rates, weighting coefficients, and fusion ratios. Such manual tuning not only increases the computational overhead but also limits the model’s adaptability and scalability across diverse and heterogeneous tasks. The optimal set of hyperparameters may vary significantly depending on task complexity, modality composition, and data scale. Consequently, the lack of an automated tuning mechanism can hinder the generalization ability of the system when deployed in unseen or dynamically changing environments.
Second, the present experimental setup assumes that each incremental task is associated with sufficient and balanced training data. However, in many real-world scenarios, data availability is often constrained — datasets may be noisy, incomplete, or heavily imbalanced across classes or modalities. This mismatch between the controlled experimental conditions and practical data constraints may reduce the effectiveness of the proposed method in realistic deployment scenarios.
Third, the current framework primarily focuses on static benchmarks such as ImageNet100, which, while valuable for controlled evaluation, do not fully capture the temporal, contextual, and distributional complexities of real-world continual learning environments. The model’s capacity to handle dynamically evolving tasks, streaming data, or asynchronous multimodal inputs remains to be systematically examined.
Lastly, although qualitative and quantitative analyses demonstrate strong feature retention and reduced catastrophic forgetting, the computational efficiency and memory footprint of the method have not been deeply explored. These aspects are critical for practical implementation, especially in edge or resource-limited environments, where continual learning systems must balance performance with real-time efficiency.
\subsection{Future Work}
Building upon the current findings, several directions can be explored to further enhance the performance and applicability of the proposed framework.
First, future research could focus on adaptive hyperparameter optimization strategies, such as reinforcement learning–based or gradient-free optimization techniques, to automatically adjust learning dynamics in response to task complexity and data variability. Incorporating such mechanisms would improve model robustness, reduce human intervention, and enable efficient scaling to large-scale or continuously evolving datasets.
Second, integrating few-shot learning and meta-learning paradigms into the continual learning pipeline offers a promising avenue for addressing data scarcity and imbalance. By leveraging meta-knowledge or transferable priors, the model could generalize effectively from limited or unevenly distributed samples while maintaining resistance to catastrophic forgetting. This integration could make the framework more applicable to real-world settings where acquiring abundant, balanced data is impractical.
Third, extending the framework to real-world multimodal applications represents an important step toward validating its practical utility. Potential domains include anomaly detection in industrial monitoring, remote sensing and satellite imagery analysis, autonomous driving perception systems, and video-based object tracking, where dynamic and continual multimodal understanding is essential. Evaluating the model in these scenarios would not only test its robustness under non-stationary conditions but also uncover domain-specific challenges and adaptation requirements.
Finally, incorporating self-supervised or online adaptation mechanisms could further enhance the system’s capability for lifelong learning. By enabling the model to self-update in response to environmental changes or unseen modalities, such extensions could lead to a truly autonomous and continuously evolving multimodal learning framework.
In summary, addressing these future directions would not only improve the scalability, adaptability, and efficiency of the proposed approach but also contribute to the broader goal of developing intelligent systems capable of sustainable, real-world continual learning.

\section{Conclusion}
\label{sec5}
This paper proposes a novel CCL framework called CalFuse, which integrates visual and textual features within a continual learning paradigm. To address the feature distortion caused by directly fine-tuning VLMs such as CLIP, CalFuse introduces FC to preserve the original semantic structure of pre-trained features. This design helps maintain the model’s generalization ability and stability across tasks. In addition, CalFuse employs a stage-specific parameter fusion strategy to mitigate catastrophic forgetting. By adaptively decomposing and fusing parameters at different task stages, the model can effectively balance the learning of new knowledge with the retention of previous knowledge. This approach enables CalFuse to achieve strong performance in both incremental accuracy and long-term knowledge retention. Experimental results on benchmarks like ImageNet100 and CIFAR-100 demonstrate that CalFuse outperforms traditional and recent CCL methods, confirming the advantages of multi-modal feature integration and the proposed training strategies. While some hyperparameters in CalFuse were manually set, future work will focus on developing adaptive parameter tuning methods. Additionally, we aim to extend CalFuse to few-shot CCL scenarios, which better reflect real-world conditions where new class data is limited.

\bibliographystyle{IEEEtran}
\bibliography{refs1}

\begin{IEEEbiography}[{\includegraphics[width=1in,height=1.25in,clip,keepaspectratio]{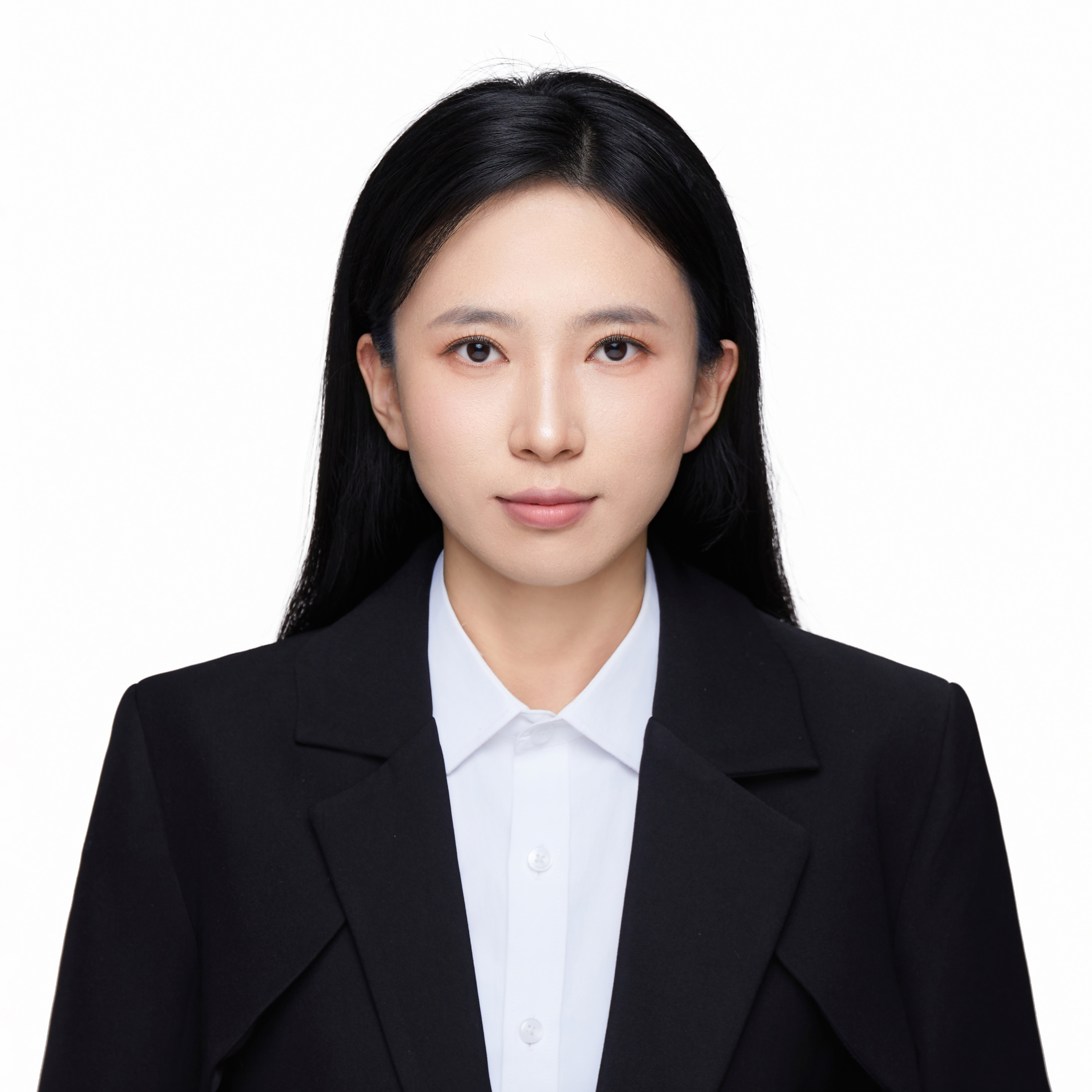}}]{Juncen Guo} received her bachelor's degree in Information and Computing Science and master's degree in Management Science and Engineering from Tianjin University of Technology in 2014 and 2018, respectively. She is working toward the Ph.D. degree in computer science with the Academy for Engineering and Technology, Fudan University, Shanghai, China. She has worked in the construction and management of enterprise big data platform for 4 years. Her research interests include computer vision and intelligent signal processing, with an emphasis on online evolutionary learning and continuous learning of multimodal large models.
\end{IEEEbiography}

\begin{IEEEbiography}[{\includegraphics[width=1in,height=1.25in,clip,keepaspectratio]{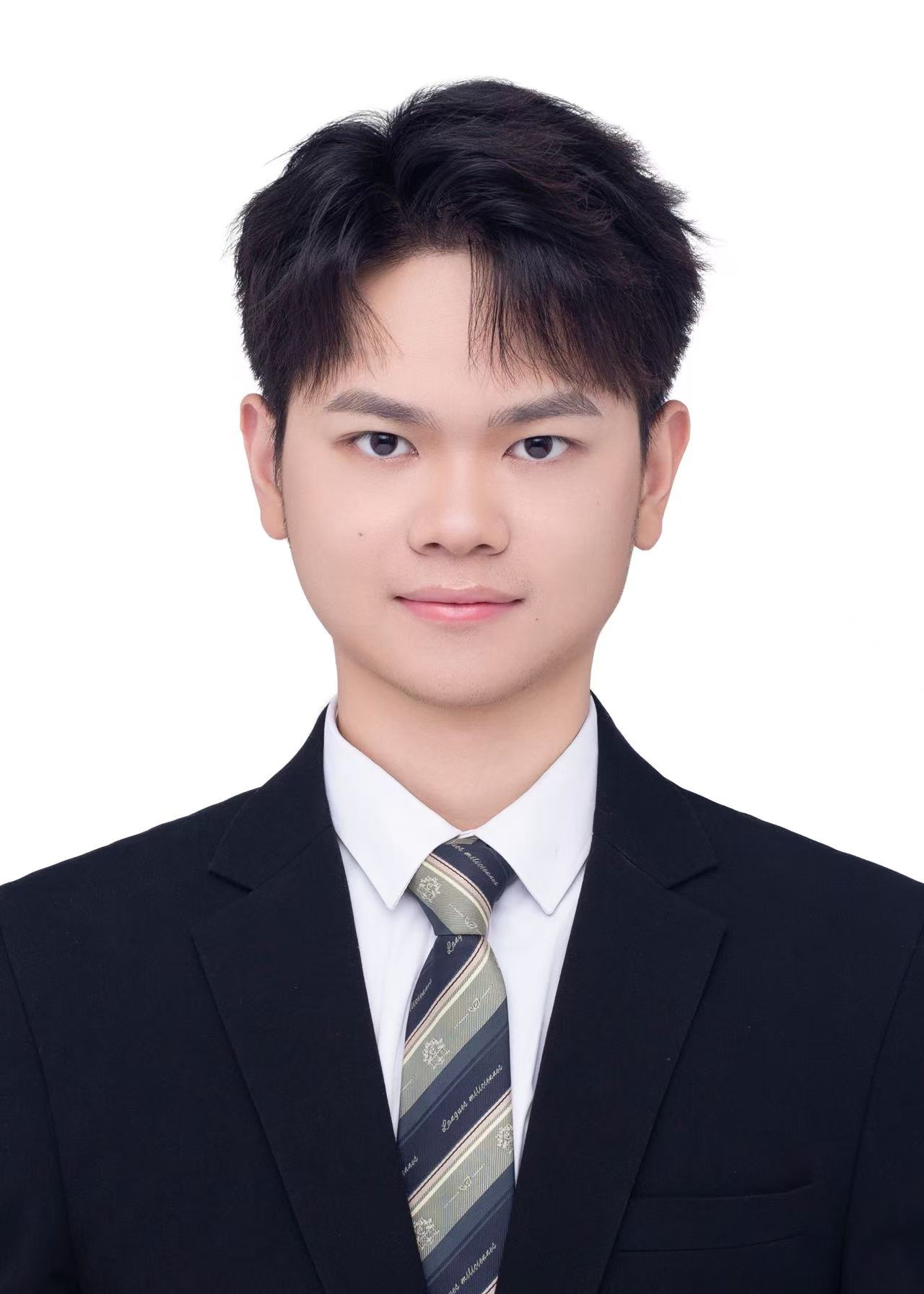}}]{Siao Liu} received the Ph.D. degree in Intelligent Robotics and Advanced Manufacturing from Fudan University, Shanghai, China, in 2025, and the B.Eng. degree in Mechanical, Electrical \& Information Engineering from Shandong University (Weihai), China, in 2020. He is currently with the School of Future Science and Engineering, Soochow University, Suzhou, China. His research interests include computer vision, long-horizon robotic learning, and reinforcement learning, with a focus on robust perception–action systems.
\end{IEEEbiography}

\begin{IEEEbiography}[{\includegraphics[width=1in,height=1.25in,clip,keepaspectratio]{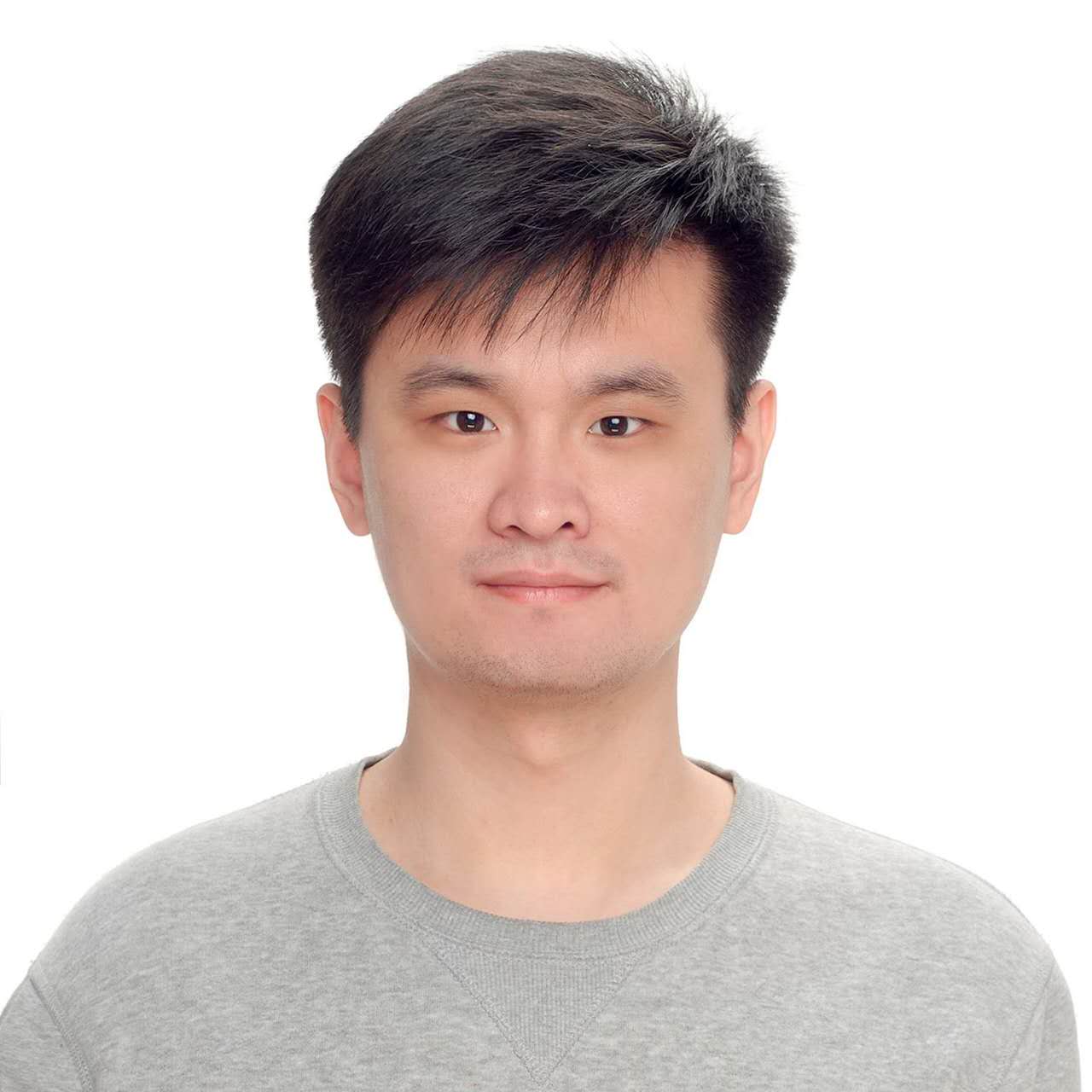}}]{XiaoguangZhu} is a postdoctoral researcher at the University of California, Davis. He completed his academic journey at Shanghai Jiao Tong University, earning his Bachelor's degree in June 2014, Master's degree in March 2017, and Doctor's degree in June 2022. His research interests include representation learning, time-series forecasting, causal inference and AI for Health.
\end{IEEEbiography}

\begin{IEEEbiography}[{\includegraphics[width=1in,height=1.25in,clip]{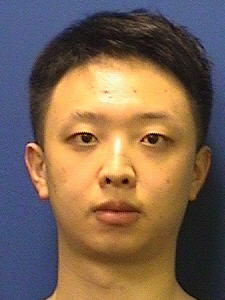}}]{Lianlong Sun} received the B.S. degree from Shanghai Jiao Tong University, Shanghai, China, and the M.S. degree from the University of Rochester, Rochester, NY, USA (2023). He is currently pursuing the Ph.D. degree in Electrical and Computer Engineering at the University of Rochester. His research interests include combinatorial optimization and dynamical systems. 
\end{IEEEbiography}

\begin{IEEEbiography}[{\includegraphics[width=1in,height=1.25in,clip,keepaspectratio]{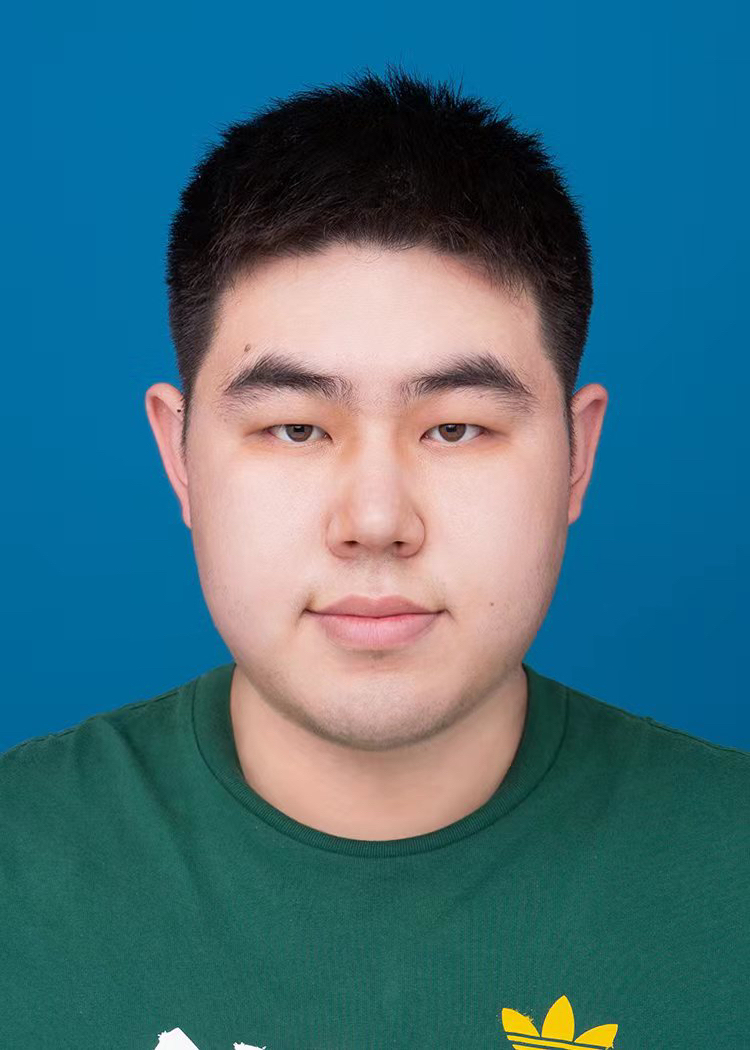}}]{Liangyu Teng} received the B.S. degree in Intelligence Science and Technology from the School of Information Science and Technology, Fudan University, Shanghai, China, in 2024. Currently, he is pursuing the M.S. degree in Computer Science and Technology with the Academy for Engineering and Technology, Fudan University. His research interests include multimodal large language models, distributed artificial intelligence, and multi-agent systems.
\end{IEEEbiography}

\begin{IEEEbiography}[{\includegraphics[width=1in,height=1.25in,clip,keepaspectratio]{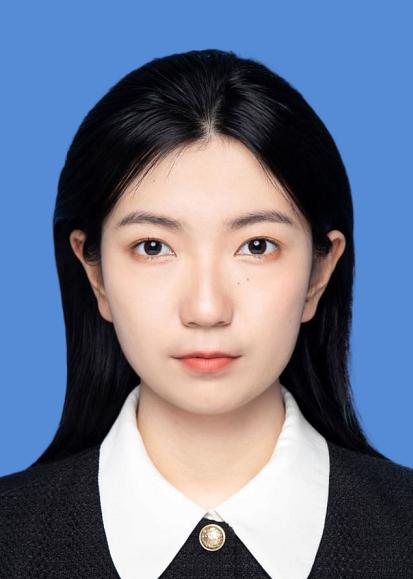}}]{Jingyi Wu} received the B.S. degree in Qingdao University, Qingdao, China, in 2023. She is currently working toward the Ph.D. degree in computer science with the Academy for Engineering and Technology, Fudan University, Shanghai, China. Her current research focuses on image processing, computer vision, and deep learning, particularly in the domains of 3D reconstrction, 3D generation, and video generation.
\end{IEEEbiography}

\begin{IEEEbiography}[{\includegraphics[width=1in,height=1.25in,clip,keepaspectratio]{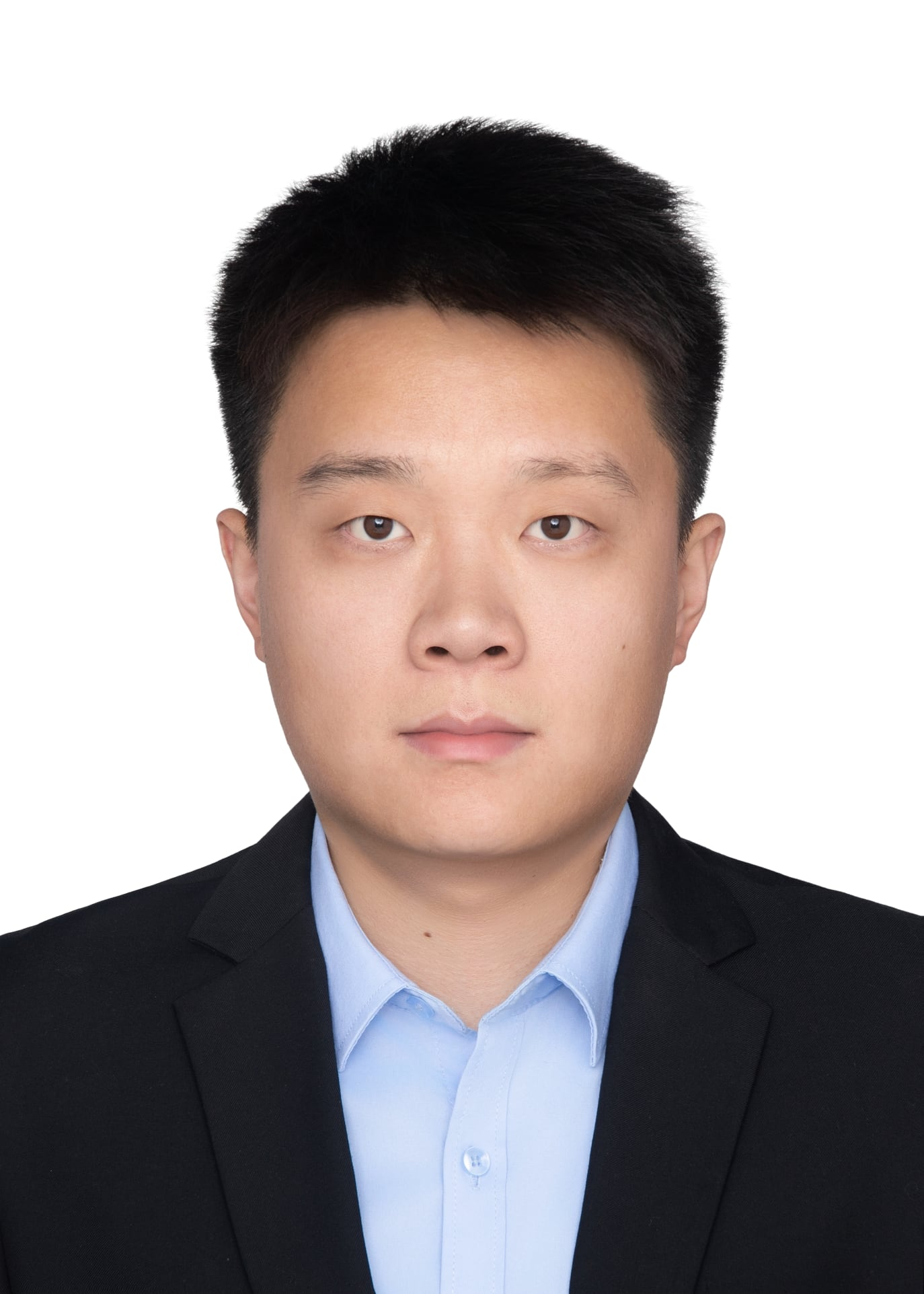}}]{Di Li} is currently a lecture at the Faculty of Electrical Engineering and Computer Science, Ningbo University, and an associate researcher in Fudan Institute on NSAI and Shanghai East-bund Research Institute on NSAI, China. He received his Ph.D. degree in Detection Technology and Automatic Equipment from Henan University of Science and Technology in 2022. His research interests include computer vision, deep learning and edge computing, with a focus on semi-supervised learning and online evolutive learning approaches.
\end{IEEEbiography}

\begin{IEEEbiography}[{\includegraphics[width=1in,height=1.25in,clip,keepaspectratio]{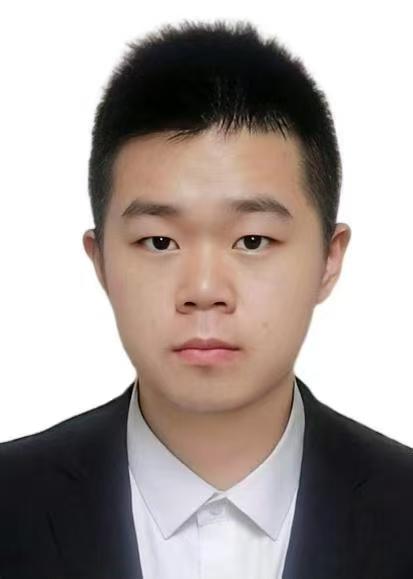}}]{Linxiao Gong} received the B.S. degree in Electronic Engineering from Fudan University, Shanghai, China, in June 2025. He is currently pursuing the Master's degree at HKUST (GZ). His research interests include LLM and data mining, focusing on the LLM-augmented data mining methodologies and agentic systems.
\end{IEEEbiography}

\begin{IEEEbiography}[{\includegraphics[width=1in,height=1.25in,clip,keepaspectratio]{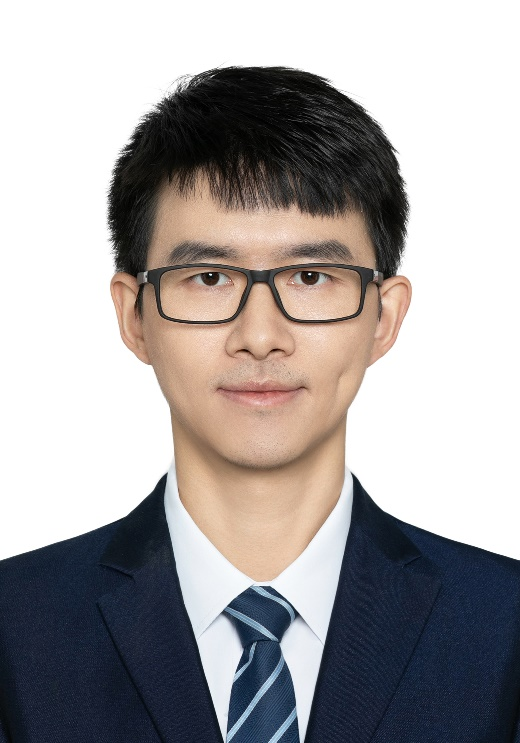}}]{Weiwei Jiang} received the B.Sc. Degree of Electronic Engineering and Ph.D. Degree of Information and Communication Engineering from the Department of Electronic Engineering, Tsinghua University, Beijing, China, in 2013 and 2018, respectively. He is currently an assistant professor with the School of Information and Communication Engineering, Beijing University of Posts and Telecommunications, and Key Laboratory of Universal Wireless Communications, Ministry of Education. His current research interests include artificial intelligence, machine learning, big data, wireless communication and edge computing. He has published more than 60 academic papers in IEEE Trans and other journals, with more than 4100 citations in Google Scholar. He is one of 2023 and 2024 Stanford's List of World's Top 2\% Scientists.
\end{IEEEbiography}
\vspace{-200pt}

\begin{IEEEbiography}[{\includegraphics[width=1in,height=1.25in,clip,keepaspectratio]{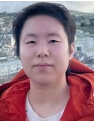}}]{Wei Zhou}(IEEE Senior Member) is an Assistant Professor at Cardiff University, United Kingdom. Previously, Wei studied and worked at other institutions such as the University of Waterloo (Canada), the National Institute of Informatics (Japan), the University of Science and Technology of China, Intel, Microsoft Research, and Alibaba Group. Dr Zhou is now an Associate Editor of IEEE Transactions on Neural Networks and Learning Systems, ACM Transactions on Multimedia Computing, Communications, and Applications (TOMM), and Pattern Recognition. Wei’s research interests span multimedia computing, perceptual image processing, and computational vision. 
\end{IEEEbiography}
\vspace{-200pt}

\begin{IEEEbiography}[{\includegraphics[width=1in,height=1.25in,clip,keepaspectratio]{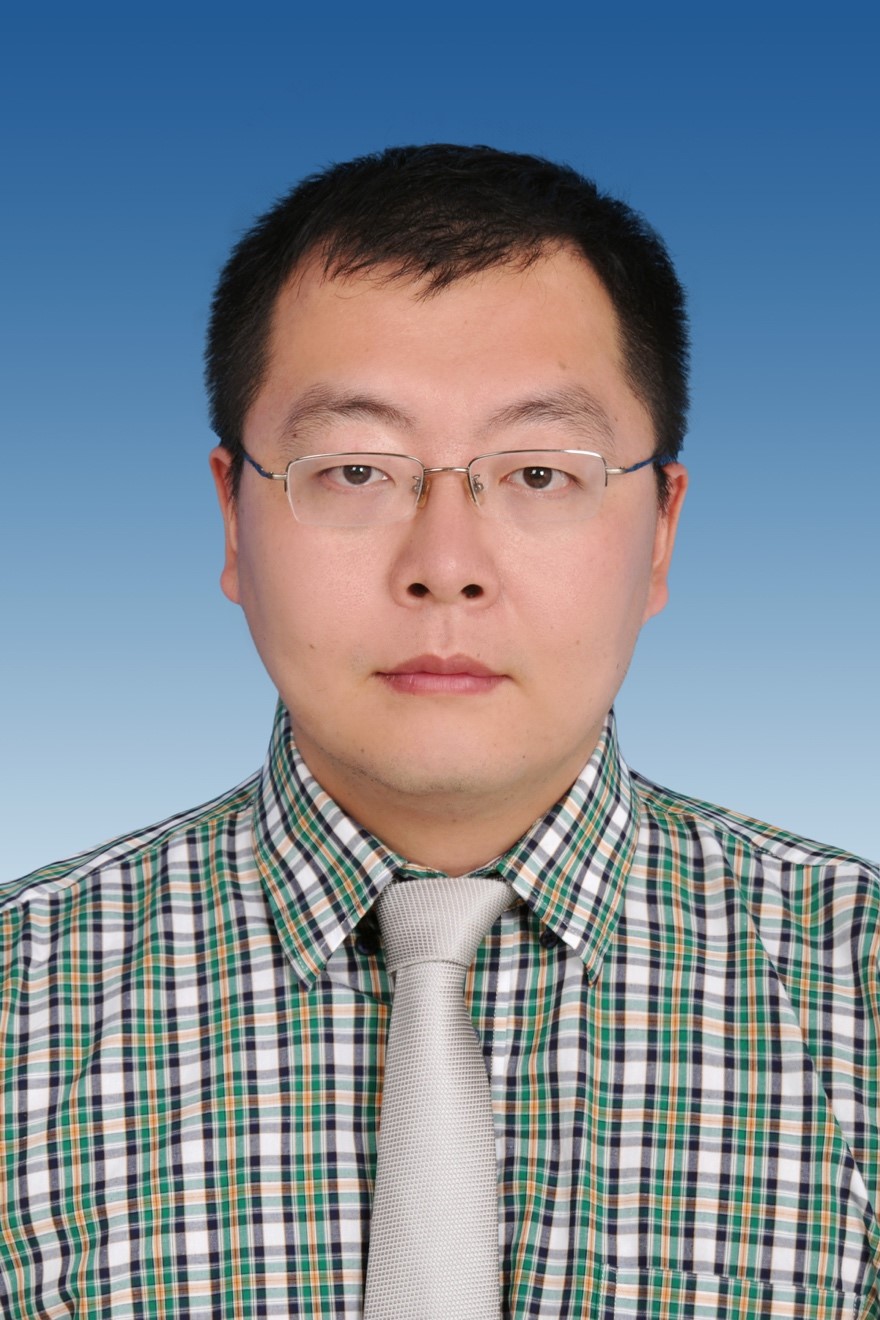}}]{Liang Song} (Senior Member, IEEE) 
  received B.E. from Shanghai Jiaotong University in 1999, M.S. from Fudan University in 2002, and Ph.D. from the University of Toronto in 2006, respectively.  He is currently a Professor with Fudan University, as the director of Fudan Institute on Networking Systems of AI (FINSAI), with more than 30 faculties. along with numerous distinguished or adjunct appointments, e.g., Chairman of Institute on Networking Systems of AI, the Professor with the University of Toronto, and the Chairman of Shanghai 5G-VR Alliance, among others. His research interests converge communication networks and AI systems, where he has published more than 200 referred papers, 8 monographs, and invented over 100 patents. 
\end{IEEEbiography}

\end{document}